\pdfoutput=1

\documentclass[11pt]{article}

\usepackage[final]{acl}

\usepackage{times}
\usepackage{latexsym}

\usepackage[T1]{fontenc}

\usepackage[utf8]{inputenc}

\usepackage{microtype}

\usepackage{inconsolata}

\usepackage{graphicx}

\usepackage{booktabs}
\usepackage{multirow}
\usepackage{makecell}
\usepackage{amsmath}
\usepackage{xcolor,colortbl}
\definecolor{tableaccent}{RGB}{46,77,104}
\newcommand{\resulttablestyle}{%
  \small
  \setlength{\tabcolsep}{3pt}%
  \renewcommand{\arraystretch}{1.10}%
}
\usepackage{float}
\usepackage{amsfonts}
\usepackage{amssymb}
\usepackage{pifont}
\usepackage{tabularx}
\usepackage{enumitem}
\usepackage{tcolorbox}
\tcbuselibrary{breakable}

\usepackage{listings}
\lstdefinestyle{json}{
    basicstyle=\footnotesize\ttfamily,
    columns=fullflexible,
    keepspaces=true,
    commentstyle=\color{gray},
    stringstyle=\color{blue},
    keywordstyle=\color{red},
    numbers=none,
    showstringspaces=false,
    breaklines=true,
    captionpos=b,
    tabsize=2,
    language=Python 
}

\usepackage{framed}
\usepackage{algpseudocode}


\title{ATOD: An Evaluation Framework and Benchmark for Agentic Task-Oriented Dialogue Systems}

\author{
    \textbf{Yifei Zhang}$^{1,2}$\thanks{Work performed during internship at Amazon.} \quad
    \textbf{Hooshang Nayyeri}$^{1}$ \quad
    \textbf{Rinat Khaziev}$^{1}$
    \\
    \textbf{Emine Yilmaz}$^{1,3}$ \quad
    \textbf{Gokhan Tur}$^{1,4}$ \quad
    \textbf{Dilek Hakkani-T\"ur}$^{1,4}$ \quad
    \textbf{Hari P Thadakamalla}$^{1}$
    \\
    $^{1}$Amazon \quad
    $^{2}$Emory University \quad
    $^{3}$University College London
    \\
    $^{4}$University of Illinois Urbana--Champaign
    \\
    \texttt{\{jimmyzyf, hooshang, rinatk, gokturt, hakkanit, thadakah\}@amazon.com} \\
    \texttt{eminey@amazon.co.uk}
}

\hypersetup{
  pdftitle={ATOD: An Evaluation Framework and Benchmark for Agentic Task-Oriented Dialogue Systems},
  pdfauthor={Yifei Zhang; Hooshang Nayyeri; Rinat Khaziev; Emine Yilmaz; Gokhan Tur; Dilek Hakkani-T\"ur; Hari P Thadakamalla},
  pdfkeywords={task-oriented dialogue, evaluation, goal tracking, agentic memory}
}

\begin{document}
\maketitle

\begin{abstract}
Agentic task-oriented dialogue (TOD) requires systems to track concurrent goals, dependencies, and long-horizon state. We examine goal-lifecycle recovery from fixed dialogue trajectories. ATOD contains 1{,}000 synthetic dialogues annotated for six advanced-TOD properties, and ATOD-Eval defines metrics for dependency-sensitive completion, memory recall, and proactivity. We implement a symbolic--vector memory evaluator for ATOD-Eval. Among five prompt-only predictors and six matched-backbone memory baselines, our evaluator is the only configuration above 90\% in both goal detection F1 and conditional status accuracy on medium dialogues. On complex dialogues, no baseline exceeds it on both metrics; it has the highest conditional status accuracy within the memory-based block and the lowest measured per-turn latency. These experiments assess lifecycle tracking rather than interactive agent task success. A cross-family judge swap and a manual audit with 94.0\% agreement provide initial checks on measurement reliability. Code and data will be released at \url{https://github.com/amazon-science/ATOD}.
\end{abstract}

\section{Introduction}

\renewcommand{\arraystretch}{1.2}
\begin{table}[ht]
\centering
\footnotesize
\setlength{\tabcolsep}{3pt}
\caption{Illustrative interaction and the capabilities it motivates. Utterances are abbreviated for readability; an annotated ATOD excerpt appears in Appendix~\ref{appendix:dialogue_example}.}
\label{tab:example_dialogue}
\begin{tabularx}{\columnwidth}{@{}>{\raggedright\arraybackslash}p{0.27\columnwidth}>{\raggedright\arraybackslash}X >{\raggedright\arraybackslash}p{0.24\columnwidth}@{}}
\toprule
\rowcolor{tableaccent!9}
\textbf{User request} & \textbf{Agent response} & \textbf{Capabilities} \\
\midrule
Book a flight and hotel. & Offers coordinated flight and hotel options. & \textcolor{tableaccent}{Multi-goal concurrency} \\
\rowcolor{tableaccent!3}
Remind me to pack. & Retains a reminder for the night before travel. & \textcolor{tableaccent}{Long-horizon memory} \\
What is the weather? & Answers a weather query while bookings remain active. & \textcolor{tableaccent}{Interleaving} \\
\rowcolor{tableaccent!3}
Book dinner after my flight. & Uses arrival time to constrain the dinner reservation. & \textcolor{tableaccent}{Dependencies} \\
Arrange an airport ride. & Schedules the ride and offers to send details later. & \textcolor{tableaccent}{Asynchronous execution; proactivity} \\
\rowcolor{tableaccent!3}
Send details closer to departure. & Retains the deferred follow-up. & \textcolor{tableaccent}{Long-horizon memory; asynchrony} \\
\bottomrule
\end{tabularx}
\end{table}

Task-oriented dialogue (TOD) has traditionally been studied as short, sequential interactions that resolve user requests turn by turn~\cite{rastogi2020towards,budzianowski2018multiwoz}. LLM-based assistants such as ChatGPT~\cite{openai2024chatgptagent} and Gemini~\cite{comanici2025gemini} have raised expectations for more complex interactions. We use \textbf{Advanced TOD} to describe settings in which agents coordinate concurrent goals, suspend and resume them, respect evolving dependencies, maintain long-horizon memory, handle asynchronous execution, and act proactively. Table~\ref{tab:example_dialogue} shows an example with interlocked flight, hotel, and restaurant bookings.

Evaluating advanced TOD is harder than slot-filling: a system can complete individual goals yet fail to manage their dependencies, or track status for detected goals while missing others. Existing benchmarks, including MultiWOZ~\cite{budzianowski2018multiwoz}, SGD~\cite{rastogi2020towards}, and $\tau$-Bench~\cite{yao2024tau}, do not provide turn-level lifecycle annotations across this combination of behaviors; TOAD~\cite{liu2024toad} and LUCID~\cite{stacey2024lucid} cover only parts of it. General LLM judges~\cite{liu2023g,zheng2023judging} and memory benchmarks such as LoCoMo~\cite{maharana2024evaluating} also do not evaluate dependency-aware goal trajectories. To our knowledge, none of the surveyed benchmarks supports fine-grained evaluation of the full capability set studied here.

This paper makes three contributions. \textbf{(i)} \textbf{ATOD}, a benchmark of 1{,}000 synthetic dialogues that jointly covers all six advanced-TOD properties at turn-level granularity, together with its full generation pipeline (\S\ref{sec:dataset_construction}). \textbf{(ii)} \textbf{ATOD-Eval}, a multi-dimensional evaluation framework with metrics for dependency-sensitive completion (dGCR), memory recall (MRA), proactivity (PE), and response quality, providing complementary offline and online signals (\S\ref{sec:protocols}). \textbf{(iii)} An \textbf{agentic memory evaluator} with a dual symbolic--vector store and a four-stage turn-level pipeline (extract / match / update-or-insert / audit) that balances goal detection and conditional status tracking on fixed complex dialogues at the lowest measured latency among matched-backbone baselines (\S\ref{sec:memory_system},~\S\ref{sec:results}).

\section{Related Work}
\label{sec:related}

\subsection{TOD Systems Evaluation}
G-Eval~\cite{liu2023g}, AlpacaEval~\cite{dubois2024length}, and MT-Bench~\cite{zheng2023judging} assess generated text or assistant response quality rather than explicitly tracking goal lifecycles. TOD evaluation spans turn-level user satisfaction~\cite{walker2000towards}, dialogue-level quality estimation~\cite{deng2022user,komma2023toward}, and LLM-based user simulation~\cite{kazi2024large,davidson2023user}. Recent benchmarks ($\tau$-Bench~\cite{yao2024tau}, AutoEval-ToD~\cite{jain2025autoeval}, TD-EVAL~\cite{acikgoz2025td}) focus on inform/success rates rather than advanced-TOD capabilities. Their slot-level or rubric-based metrics are not directly comparable to our turn-level lifecycle metrics, so we discuss them as related settings rather than direct baselines. $\tau^2$-Bench~\cite{barres2025tau2} evaluates conversational agents in a shared environment where both the agent and user can act through tools. ATOD instead evaluates recovery of turn-level goal lifecycles from fixed dialogues; it does not replace interactive evaluation.

\subsection{TOD Datasets and Benchmarks}
MultiWOZ~\cite{budzianowski2018multiwoz}, SGD~\cite{rastogi2020towards}, RADDLE~\cite{peng2020raddle}, $\tau$-Bench~\cite{yao2024tau}, and MS-TOD~\cite{du2025bridging} are human-curated TOD datasets that support dialogue state tracking and task completion but are largely single-session. Synthetic datasets (TOPDIAL~\cite{wang2023target}, TOAD~\cite{liu2024toad}, LUCID~\cite{stacey2024lucid}, SynthDST~\cite{kulkarni2024synthdst}) add personalization or proactivity but lack joint coverage. In broader agent evaluation, AgentBench~\cite{liu2024agentbench}, T1~\cite{chakraborty2026t}, and AgentChangeBench~\cite{rana2025agentchangebench} target environments, tool-use planning, and goal-shift adaptation, but none provides turn-level goal lifecycle annotations with dependency-aware metrics.

\subsection{Memory for Dialogue Systems}
RAG~\cite{lewis2020retrieval}, MemoChat~\cite{lu2023memochat}, and MemoryBank~\cite{zhong2024memorybank} support session-level recall; MemGPT~\cite{packer2023memgpt}, A-Mem~\cite{xu2025mem}, Mem0~\cite{chhikara2025mem0}, and MemOS~\cite{li2025memos} introduce structured long-term memory; Chronos~\cite{sen2026chronos} and SGMem~\cite{wu2025sgmem} use event-based representations; HierTOD~\cite{mo2024hiertod} and TACT~\cite{yoon2025beyond} target hierarchical and proactive TOD. Benchmarks like LoCoMo~\cite{maharana2024evaluating} and THEANINE~\cite{ong2024towards} study memory temporality but not dependency-aware goal evaluation, which ATOD complements.

\begin{figure*}[ht]
    \centering
    \includegraphics[width=0.99\linewidth]{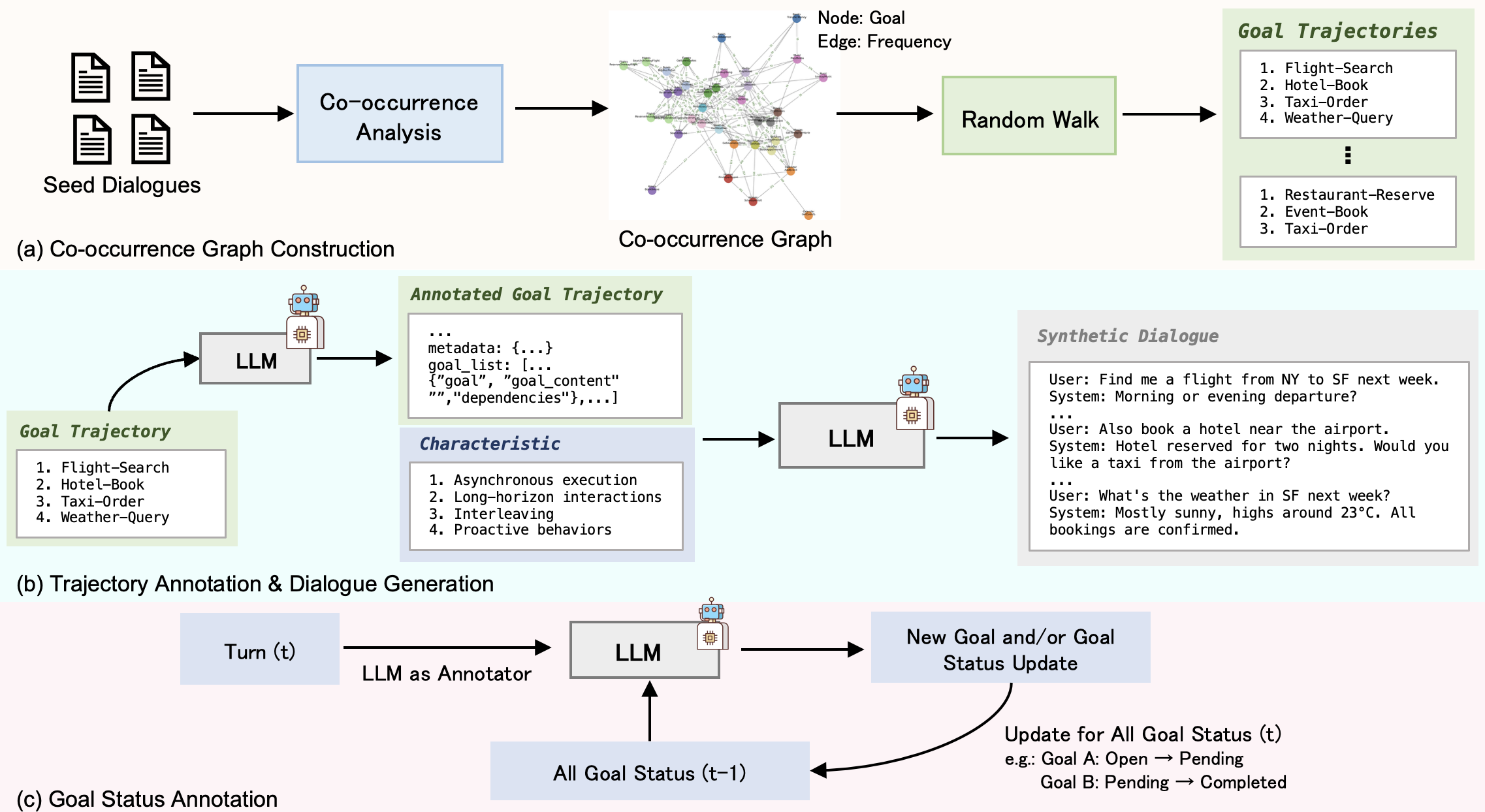}
    \caption{ATOD dataset curation pipeline. (a) \textbf{Co-occurrence Graph \& Trajectory Sampling}: build a goal co-occurrence graph from seed dialogues and sample diverse multi-goal trajectories via random walks; (b) \textbf{Trajectory Annotation \& Dialogue Generation}: annotate slot values, dependencies, and complexity, then generate dialogues conditioned on trajectories; (c) \textbf{Goal Status Annotation}: label active goals and lifecycle states at each turn for fine-grained tracking of dialogue progress.}
    \label{fig:synthetic_pipeline}
\end{figure*}

\section{Problem Formulation}
\label{sec:problem}

\subsection{Characteristics of Advanced TOD}
\label{sec:advanced_tod}

Prior TOD benchmarks~\cite{budzianowski2018multiwoz,rastogi2020towards,yao2024tau} typically isolate one or two of the behaviors below, while realistic agentic interactions exhibit them jointly. We therefore characterize advanced TOD by six properties that any evaluation must support together: \emph{multi-goal concurrency}, where users pursue multiple objectives in parallel; \emph{interleaving}, where goals are suspended and resumed rather than confined to contiguous turn spans; \emph{dependency management}, where goals carry prerequisite or blocking relations (e.g., \textsc{Payment} depends on \textsc{Booking}); \emph{long-horizon memory}, where state is preserved across many turns and goal interruptions; \emph{asynchronous execution}, where delayed goals remain \textsc{Pending} until external conditions hold, including system-initiated follow-ups; and \emph{proactivity}, where the system offers reminders or suggestions without explicit prompting.

\subsection{Task Formulation}

We formalize advanced TOD as a corpus $\mathcal{D} = \{(\mathcal{G}_i, \mathcal{C}_i)\}_{i=1}^N$, where $\mathcal{C}_i = \{c_{i,t}\}_{t=1}^{T_i}$ is the turn sequence of dialogue $i$ and $\mathcal{G}_i$ is its goal set with explicit inter-goal dependencies. In traditional TOD a goal occupies a contiguous span of turns, whereas an advanced-TOD goal $g \in \mathcal{G}_i$ may span disjoint intervals of $\mathcal{C}_i$ as it is initiated, suspended, and resumed. We therefore represent each goal by its turn-aligned \textit{status trajectory} $\{\mathrm{Status}(g,t)\}_{t=1}^{T_i}$ (e.g., \textsc{Open}$\to$\textsc{Pending}$\to$\textsc{Completed}), capturing a non-contiguous lifecycle that prior single-span representations cannot. The evaluation objective is to measure how well a system maintains and updates these trajectories in both \textit{offline} (over completed dialogues) and \textit{online} (turn by turn) settings.

\section{ATOD: A Synthetic Dialogue Dataset and Generation Pipeline}
\label{sec:dataset_construction}

\textbf{ATOD} is a synthetic dataset and generation pipeline for benchmarking the advanced-TOD characteristics in \S\ref{sec:advanced_tod}. It contains 1{,}000 annotated multi-turn dialogues (428 medium, 572 complex) spanning 16 domains and 36 intents, with an average of 26.6 turns and 6.5 goals per dialogue. Construction includes staged quality checks (Figure~\ref{fig:synthetic_pipeline}): a verifier pass filters or regenerates inconsistent instances, and an author manually screens the generated dialogues and annotations for evident defects. Appendix~\ref{appendix:quality_control} documents these checks, and Appendix~\ref{appendix:dataset_quality} reports automated ratings. We plan to release ATOD as a zero-shot, test-only benchmark.

\subsection{Co-occurrence Graph Construction and Goal Trajectory Sampling}
\label{sec:graph_construction}

Independent goal sampling yields implausible combinations, while fixed templates limit diversity. We therefore construct a goal co-occurrence graph $G=(V,E)$ from a seed dialogue dataset, where nodes are goals and weighted edges capture empirical co-occurrence frequency. Candidate goal sets $S=\{g_1,\ldots,g_k\}$ are sampled by stratified random walks of varying lengths over $G$ (Figure~\ref{fig:synthetic_pipeline}(a)), preserving common goal combinations while adding variation. The procedure is dataset-agnostic; we instantiate it on Schema-Guided Dialogue (SGD)~\cite{rastogi2020towards}, producing multi-domain trajectories with interleaved, long-horizon goals.

\subsection{Annotation of Goal Trajectories and Complexity Categorization}
\label{sec:annotation_classification}

Each sampled set $S$ is instantiated into a concrete trajectory via Claude-Sonnet-4 annotation (Figure~\ref{fig:synthetic_pipeline}(b)), yielding slot values, inter-goal dependencies $D_S$ (e.g., prerequisites), and natural-language descriptions. Trajectories are tagged with a complexity label $c(S)$ derived from quantitative attributes (goal count, dependency density) and qualitative factors (interleaving, proactivity opportunities); see Appendix~\ref{appendix:complexity_criteria}. Quality control combines duplicate filtering, retries, and a dedicated verifier judgment of trajectory validity and generated-dialogue consistency (Appendix~\ref{appendix:quality_control}).

\subsection{Dialogue Generation}
\label{sec:dialog_generation}

Claude-Sonnet-4 synthesizes each dialogue conditioned on the annotated trajectory $\tau=(S,D_S,c(S))$. Goals, dependencies, and complexity are combined into structured prompts (Appendix~\ref{appendix:synthetic_gen_annotation_prompt}) that produce a multi-turn conversation $\mathcal{C}=\{c_t\}_{t=1}^T$ with interleaving, asynchrony, proactive assistance, and dependency-aware coordination. This conditioning is central: the generator sees the global trajectory, but the evaluated system must recover its dynamics from dialogue evidence alone.

\subsection{Turn-level Goal Status Annotation}
\label{sec:status_annotation}

A final Claude-Sonnet-4 pass labels each goal at every turn (Figure~\ref{fig:synthetic_pipeline}(c)) with an active goal set $\mathcal{A}_t \subseteq S$ and statuses $\mathrm{Status}(g,t) \in \{$\textsc{Not\_Mentioned}, \textsc{Open}, \textsc{Pending}, \textsc{Completed}, \textsc{Failed}, \textsc{Abandoned}$\}$. The resulting turn-aligned \texttt{status\_history} supports evaluation of interleaved, asynchronous, and multi-goal trajectories.

\begin{table*}[ht]
\centering
\renewcommand{\arraystretch}{1.15}
\caption{Selected operational dimensions of representative TOD benchmarks; a coverage summary, not a quality comparison. ``Goal Status Anno.'' marks explicit per-turn lifecycle labels. \ding{51}: present, \ding{55}: absent.}
\label{tab:dataset_comparison}
\resizebox{0.85\textwidth}{!}{
\begin{tabular}{lcccccc}
\toprule
\textbf{Dataset}                          & \textbf{Avg. Turns} & \textbf{Async.} & \textbf{Dependency} & \textbf{Interleaving} & \textbf{Proactivity} & \textbf{Goal Status Anno.} \\
\midrule
MultiWOZ~\cite{budzianowski2018multiwoz}  & 13                  & \ding{55}      & \ding{55}           & \ding{55}             & \ding{55}          & \ding{55} \\
SGD~\cite{rastogi2020towards}             & 20                  & \ding{55}      & \ding{55}           & \ding{55}             & \ding{55}          & \ding{55} \\
TOPDIAL~\cite{wang2023target}             & 12                  & \ding{55}      & \ding{55}           & \ding{55}             & \ding{51}          & \ding{55} \\
MS-TOD~\cite{du2025bridging}              & 7                   & \ding{55}      & \ding{51}           & \ding{51}             & \ding{55}          & \ding{51} \\
TOAD~\cite{liu2024toad}                   & 5                   & \ding{55}      & \ding{51}           & \ding{51}             & \ding{51}          & \ding{55} \\
LUCID~\cite{stacey2024lucid}              & 21                  & \ding{51}      & \ding{51}           & \ding{55}             & \ding{55}          & \ding{55} \\
\midrule
\rowcolor{blue!6}
\textbf{ATOD (Ours)}                     & \textbf{27}         & \ding{51}      & \ding{51}           & \ding{51}             & \ding{51}          & \ding{51} \\
\bottomrule
\end{tabular}
}
\end{table*}

\subsection{Dataset Coverage}
\label{sec:dataset_coverage}

ATOD spans diverse domains and goal complexities, from two-goal cases to interdependent long-horizon workflows. Table~\ref{tab:dataset_comparison} highlights the operational dimensions for which comparable annotations or behaviors are identifiable in published benchmark descriptions; ATOD combines these with concurrency and long-horizon goal trajectories. We use the full collection as a held-out benchmark and will release the dataset, generation pipeline, evaluation harness, and prompt templates upon publication. Table~\ref{tab:dataset_statistics} reports the evaluation split.

\begin{table}[ht]
\centering
\footnotesize
\setlength{\tabcolsep}{3pt}
\renewcommand{\arraystretch}{1.08}
\caption{ATOD statistics by dialogue complexity.}
\label{tab:dataset_statistics}
\begin{tabular*}{\columnwidth}{@{\extracolsep{\fill}}lccc@{}}
\toprule
\textbf{Statistic} & \textbf{Medium} & \textbf{Complex} & \textbf{Overall} \\
\midrule
Dialogues          & 428     & 572     & 1,000 \\
Avg. turns         & 17.4    & 33.5    & 26.6 \\
Avg. goals         & 3.7     & 8.5     & 6.5 \\
With dependencies  & 18.7\%  & 73.4\%  & 50.0\% \\
\bottomrule
\end{tabular*}
\end{table}

\section{Agentic Memory System}
\label{sec:memory_system}

\begin{figure*}[ht]
\centering
\includegraphics[width=0.97\linewidth]{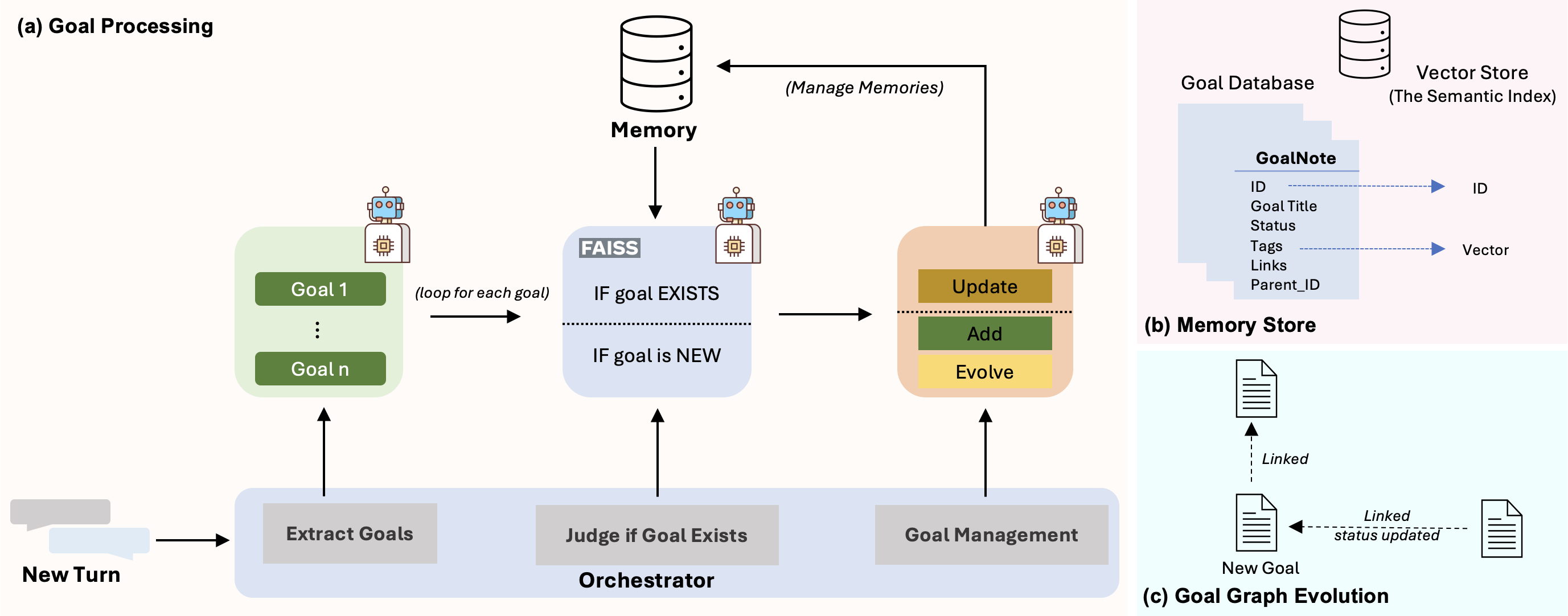}
\caption{Architecture of the agentic memory system. (a) Turn-level goal extraction, existence checking, updating/inserting, and proactive auditing. (b) Dual memory store with symbolic metadata and semantic embeddings. (c) Dependency graph evolution for newly inserted goals, with explicit links and status transitions.}
\label{fig:agentic_memory_system}
\end{figure*}

We instantiate ATOD-Eval with an \textit{agentic memory system} (Figure~\ref{fig:agentic_memory_system}) that maintains and updates goal trajectories from dialogue text alone. An LLM-only approach re-reads the full history at every turn, while flat memory does not preserve goal structure. Our system instead uses a symbolic store for lifecycle states and dependencies, together with a vector store for retrieval under lexical variation (\S\ref{sec:memory_store}). A turn-level pipeline updates both stores (\S\ref{sec:turn_processing}).

\subsection{Dual Memory Store}
\label{sec:memory_store}

As shown in Figure~\ref{fig:agentic_memory_system}(b), the memory system maintains (i) a \emph{structured goal database} $\mathcal{D}_{\mathrm{sym}}$ for state and dependency metadata, and (ii) a \emph{semantic vector store} $\mathcal{D}_{\mathrm{vec}}$ (e.g., FAISS~\cite{douze2024faiss}) for similarity-based retrieval. Each goal stores an identifier, lifecycle state, turn-level status history, textual description, dependency links, and embedding. The symbolic store preserves exact state and dependency information, while the vector store retrieves related goals expressed in different words.

\subsection{Turn-Level Processing Pipeline}
\label{sec:turn_processing}

At each turn $t$, given utterance $u_t$ and context $c_t$, the system extracts candidate goals $\mathcal{G}_t=\mathrm{extract}(u_t,c_t)$. Each candidate $g_t^{(i)}$ is then matched to an existing goal and updated, or inserted as a new goal and evolved in the dependency graph; a final auditing step keeps active states consistent:
\[
\begin{gathered}
(u_t,c_t) \xrightarrow{\mathrm{extract}} \mathcal{G}_t, \\[-0.15em]
g_t^{(i)} \xrightarrow{\mathrm{match}}
\begin{cases}
\mathrm{update}(g^\ast), & \mathrm{Match}=1, \\
\mathrm{insert{+}evolve}(g_t^{(i)}), & \mathrm{Match}=0.
\end{cases}
\end{gathered}
\]

\noindent\emph{Stage 1: existence checking.}
For each candidate $g_t^{(i)}$, we retrieve top-$k$ neighbors from $\mathcal{D}_{\mathrm{vec}}$ above a cosine similarity threshold $\tau$, then ask an LLM judge whether any neighbor is semantically equivalent. This two-stage design favors recall in retrieval and precision in verification: $\tau$ keeps unrelated goals out of the LLM call, while the judge filters surface-level matches that share keywords but differ in intent. Matched candidates go to Stage~2; novel ones to Stage~3.

\noindent\emph{Stage 2: update.}
On a match, the lifecycle is advanced (e.g., \textsc{Pending}$\to$\textsc{Completed}), slot values are refreshed, and dependencies are preserved.

\noindent\emph{Stage 3: insert and evolve.}
On no match, the goal is inserted into $\mathcal{D}_{\mathrm{sym}}$ and $\mathcal{D}_{\mathrm{vec}}$, and an LLM classifies its relation to each top-$k$ neighbor as a soft \emph{link}, a hard \emph{dependency}, or unrelated, updating the dependency graph. This is essential because new goals in advanced TOD are often conditioned on existing ones.

\noindent\emph{Stage 4: proactive auditing.}
At each turn, an auditing process inspects \textsc{Open}/\textsc{Pending} goals against the latest context and triggers valid transitions, preventing stale states for asynchronously evolving goals whose status may change without explicit user restatement.

These four stages maintain dependency-aware goal states throughout the dialogue. ATOD-Eval computes its metrics from these states (\S\ref{sec:protocols}).

\section{Evaluation Metrics and Framework}
\label{sec:protocols}
ATOD-Eval covers three dimensions: task completion and efficiency (\S\ref{sec:task_completion}), agentic capability (\S\ref{sec:agentic_capability}), and response quality (\S\ref{sec:response_quality}). Task completion alone does not show whether a system recalls earlier constraints or acts proactively, while conversational quality does not show whether dependencies were handled correctly. The framework supports both offline and online evaluation.

\subsection{Task Completion and Efficiency}
\label{sec:task_completion}
We evaluate whether goals are accomplished and how efficiently they progress through the dialogue.

\noindent\emph{Dependency-Aware Goal Completion Rate (dGCR).}
Conventional completion metrics penalize systems for goals that remain unresolved even when blocked by unmet prerequisites. We therefore evaluate only \emph{decided} goals (terminal status Completed or Failed). Let $\mathcal{U}_{\mathrm{dec}}$ denote reconstructed goals whose final status in $\mathcal{D}_{\mathrm{sym}}$ is Completed or Failed, and let $\mathcal{U}_{\mathrm{comp}}\subseteq\mathcal{U}_{\mathrm{dec}}$ contain those ending in Completed. Then
\[
\mathrm{dGCR}=\frac{|\mathcal{U}_{\mathrm{comp}}|}
{|\mathcal{U}_{\mathrm{dec}}|}.
\]
Dependency sensitivity enters through the lifecycle assignment: goals blocked by unmet prerequisites remain non-terminal and therefore do not enter $\mathcal{U}_{\mathrm{dec}}$. Abandoned goals are also excluded because a cancellation is not a system failure. For example, given $g_1$ completed, $g_2$ failed, and $g_3$ pending because it is blocked by $g_2$, the denominator is $\{g_1,g_2\}$ and $\mathrm{dGCR}=1/2$. Because dGCR is computed over reconstructed goals, we read it alongside Goal Detection F1 and Status Tracking Accuracy (Table~\ref{tab:main_results}), which compare against ground truth.

\noindent\emph{Turns to Completion (NTC).}
NTC is the average number of turns between goal initiation and a decided terminal status (Completed or Failed), using the same decided-goal set as dGCR; abandoned and unresolved goals are excluded. It complements dGCR with an efficiency signal and should be read relative to dialogue length. Table~\ref{tab:metric_results} shows a larger NTC on complex dialogues, where goals are more often suspended and resumed.

\begin{table*}[ht]
\centering
\resulttablestyle
\setlength{\tabcolsep}{5pt}
\caption{Goal detection F1 and conditional status accuracy (\%) by dialogue complexity. Status accuracy is measured over detected goals; interpret jointly with detection quality. \colorbox{gray!8}{Shaded rows} indicate memory-based methods sharing the same backbone LLM. \textbf{Bold}: our method; \underline{underline}: best baseline in each column.}
\label{tab:main_results}
\begin{tabular*}{\textwidth}{@{\extracolsep{\fill}}llcccc@{}}
\toprule
\multirow{2}{*}{\textbf{Category}}
& \multirow{2}{*}{\textbf{Method}}
& \multicolumn{2}{c}{\textbf{Medium}}
& \multicolumn{2}{c}{\textbf{Complex}} \\
\cmidrule(lr){3-4} \cmidrule(lr){5-6}
& & Goal Det. F1 & Cond. Status Acc. & Goal Det. F1 & Cond. Status Acc. \\
\midrule
\multirow{5}{*}{LLM-based}
 & DeepSeek-V3.2     & 52.12 & 81.72 & 45.96 & 65.16 \\
 & Kimi-K2.5         & 66.70 & 92.90 & 66.90 & 82.80 \\
 & MiniMax-M2.5      & 72.20 & 90.20 & 66.60 & 83.90 \\
 & Mistral-Large-3   & 60.51 & 89.85 & 55.73 & \underline{85.98} \\
 & GLM-5             & 71.42 & 93.52 & 64.84 & 83.42 \\
 \midrule
\multirow{6}{*}{Memory-based}
 & \cellcolor{gray!8}Mem0~\cite{chhikara2025mem0}           & \cellcolor{gray!8}76.80 & \cellcolor{gray!8}75.60 & \cellcolor{gray!8}61.10 & \cellcolor{gray!8}71.20 \\
 & \cellcolor{gray!8}MemoChat~\cite{lu2023memochat}         & \cellcolor{gray!8}80.38 & \cellcolor{gray!8}73.88 & \cellcolor{gray!8}58.07 & \cellcolor{gray!8}66.83 \\
 & \cellcolor{gray!8}MemoryBank~\cite{zhong2024memorybank}  & \cellcolor{gray!8}82.56 & \cellcolor{gray!8}94.23 & \cellcolor{gray!8}76.86 & \cellcolor{gray!8}78.50 \\
 & \cellcolor{gray!8}RAG~\cite{lewis2020retrieval}          & \cellcolor{gray!8}85.59 & \cellcolor{gray!8}\underline{94.85} & \cellcolor{gray!8}87.13 & \cellcolor{gray!8}77.83 \\
 & \cellcolor{gray!8}A-Mem~\cite{xu2025mem}                 & \cellcolor{gray!8}87.30 & \cellcolor{gray!8}89.50 & \cellcolor{gray!8}62.00 & \cellcolor{gray!8}80.60 \\
 & \cellcolor{gray!8}LLM-Rsum~\cite{wang2025recursively}    & \cellcolor{gray!8}\underline{93.83} & \cellcolor{gray!8}89.47 & \cellcolor{gray!8}\underline{89.13} & \cellcolor{gray!8}69.95 \\
\midrule
 & \cellcolor{blue!6}\textbf{Ours} & \cellcolor{blue!6}\textbf{91.92} & \cellcolor{blue!6}\textbf{92.31} & \cellcolor{blue!6}\textbf{86.49} & \cellcolor{blue!6}\textbf{84.28} \\
\bottomrule
\end{tabular*}
\end{table*}

\subsection{Agentic Capability Metrics}
\label{sec:agentic_capability}
Beyond task success, we assess agentic behaviors such as memory recall and proactivity.

\noindent\emph{Memory Recall Accuracy (MRA).}
At each turn where a goal status changes, we construct a retrieval query $q_t$ from the affected goal description and recent context using a fixed template (Appendix~\ref{appendix:evaluation_prompt_templates}). For each $q_t \in \mathcal{Q}$, an LLM judge compares the system retrieval $\hat{M}_t$ against a ground-truth memory snapshot $M_t$ derived from the annotation and returns $r_t \in \{0,1\}$ based on sufficiency and semantic consistency. We compute $\mathrm{MRA}=|\mathcal{Q}|^{-1}\sum_{q_t\in\mathcal{Q}}r_t$, favoring useful memory over surface-form overlap.

\noindent\emph{Proactivity Effectiveness (PE).}
A response is flagged as proactive if it references a goal not in the current user turn but present in the tracked state (e.g., a reminder about a pending booking). For each such response $p_t \in \mathcal{P}$, an LLM judge returns separate binary decisions $g_t$ and $h_t$ for whether the action is \emph{grounded} (supported by context) and \emph{beneficial} (advances an unresolved goal without redundancy). We set $b_t=g_t h_t$, so a response receives credit only when both conditions hold, and compute $\mathrm{PE}=|\mathcal{P}|^{-1}\sum_{p_t\in\mathcal{P}}b_t$. Judge templates are in Appendix~\ref{appendix:evaluation_prompt_templates}.

\subsection{Response Quality Metrics}
\label{sec:response_quality}
Following~\citet{liu2023g,zheng2023judging}, \textit{turn-level relevance} scores each response against the local context, current request, and tracked goal state (averaged over turns), while \textit{dialogue-level coherence} scores the full conversation for consistency and natural progression across interleaved goals. The judge returns both scores on a 1--5 scale. We report turn-level relevance as a fraction of the maximum score ($s/5$) and retain the native 1--5 scale for dialogue-level coherence. These measures show whether improved goal management comes at a cost to interaction quality. Templates appear in Appendix~\ref{appendix:evaluation_prompt_templates}.

\subsection{Evaluation Framework}
\label{sec:evaluation_framework}
ATOD-Eval is evaluator-agnostic: any system producing per-turn goal states can be scored. In our experiments, \textbf{offline} evaluation uses completed synthetic dialogues, while \textbf{online} evaluation processes those same fixed dialogues turn by turn. Online denotes incremental lifecycle tracking, not interaction with a user or tool environment. The framework could support monitoring of deployed agents, but end-to-end agent rollouts and real-world task success are not evaluated here.

\section{Experimental Setup}
\label{sec:experimental_setup}

Evaluation covers capability, metric validity, and efficiency. We report goal detection F1 and status tracking accuracy at the final state and across normalized dialogue progress, Pearson's $r$ and Spearman's $\rho$ between dGCR and the other metrics, and per-turn latency and token usage. Implementation details appear in Appendix~\ref{appendix:memory_implement_details}.

\noindent\emph{Baselines.} All baselines are evaluated zero-shot on the same 1{,}000-dialogue set. The prompt-only judges are DeepSeek-V3.2~\cite{liu2025deepseek}, Kimi-K2.5~\cite{team2026kimi}, MiniMax-M2.5, Mistral-Large-3, and GLM-5~\cite{zeng2026glm}; each infers goal status from the full dialogue context without external memory. We also adapt Mem0~\cite{chhikara2025mem0}, MemoChat~\cite{lu2023memochat}, MemoryBank~\cite{zhong2024memorybank}, RAG~\cite{lewis2020retrieval}, A-Mem~\cite{xu2025mem}, and LLM-Rsum~\cite{wang2025recursively} to our goal-status schema by mapping their memory units to our goal representation.

\noindent\emph{Architecture-isolated comparison.} Every memory-based method, including ours, uses the same backbone LLM and the same \texttt{all-MiniLM-L6-v2} embedding model; only the memory architecture varies. We use Claude-Sonnet-4 as the backbone for all memory-based evaluators (details in Appendix~\ref{appendix:memory_implement_details}). The prompt-only judges are the LLMs under test and indicate what prompting alone achieves across model scales. No shared external judge scores these rows. The existence threshold and top-$k$ for our evaluator were selected on a disjoint 50-dialogue development set and frozen before test evaluation; we do not claim equal hyperparameter tuning across methods.

\noindent\emph{Run variability.} The reported memory experiments use temperature $0$ and seed $42$. These settings do not guarantee identical outputs from API-served models or quantify uncertainty over sampled dialogues. We treat the reported margins as point estimates, not tests of statistical significance.

\noindent\emph{Mitigating LLM circularity.} Because the evaluator and the metric judges may share a model family, we provide two checks: a cross-family judge swap with Qwen3-32B~\cite{yang2025qwen3} and a manual audit of judge decisions on a stratified subset (\S\ref{sec:judge_validity}).

\section{Results}
\label{sec:results}

\subsection{Evaluation of the Memory System}

Table~\ref{tab:main_results} reports goal detection F1 and status tracking accuracy by complexity. The two columns must be read jointly: status accuracy is computed only over detected goals, so a system can miss many goals yet still look competitive on the second column. DeepSeek-V3.2 (52.12/81.72 on medium) is the clearest case. Mistral-Large-3 (55.73/85.98 on complex) and RAG (85.59/94.85 on medium) show the same pattern more subtly. Among the evaluated configurations, only our evaluator exceeds 90\% on both axes for medium dialogues. On complex dialogues, no baseline exceeds ours (86.49/84.28) on both metrics. Mistral-Large-3 has higher conditional status accuracy (85.98), while LLM-Rsum and RAG have higher detection F1. Our complex conditional status accuracy is highest within the matched-backbone memory block, exceeding A-Mem by 3.68 points. Our claim concerns the joint result; several baselines lead in individual columns.

\noindent\emph{Architecture-isolated block.} All memory-based methods share the same backbone and embedding (\S\ref{sec:experimental_setup}), so this block compares memory architectures. LLM-Rsum obtains higher complex detection F1 than ours (89.13 vs.\ 86.49) but lower complex status accuracy (69.95 vs.\ 84.28); A-Mem and Mem0 also drop sharply in complex detection (87.30$\to$62.00 and 76.80$\to$61.10). This result is compatible with our design rationale, but component-level causal attribution requires further ablations.

\noindent\emph{Capacity reference: prompt-only judges.} Among the five prompt-only models in Table~\ref{tab:main_results}, the strongest F1 trails the best memory-based F1 by more than 20 points in each split. Table~\ref{tab:backbone_ablation} separately controls for backbone capacity by comparing our evaluator with its prompt-only backbone.

\begin{table}[ht]
\centering
\resulttablestyle
\caption{Same-backbone comparison between prompt-only judging and our agentic memory evaluator.}
\label{tab:backbone_ablation}
\begin{tabular*}{\columnwidth}{@{\extracolsep{\fill}}lcccc@{}}
\toprule
\multirow{2}{*}{\textbf{Method}} & \multicolumn{2}{c}{\textbf{Medium}} & \multicolumn{2}{c}{\textbf{Complex}} \\
\cmidrule(lr){2-3} \cmidrule(lr){4-5}
& F1 & Acc. & F1 & Acc. \\
\midrule
Prompt-only & 78.95 & 93.26 & 75.58 & 84.26 \\
\textbf{Ours} & \textbf{91.92} & \textbf{92.31} & \textbf{86.49} & \textbf{84.28} \\
$\Delta$ & $+$12.97 & $-$0.95 & $+$10.91 & $+$0.02 \\
\bottomrule
\end{tabular*}
\end{table}

\begin{figure}[ht]
    \centering
    \includegraphics[width=0.95\columnwidth]{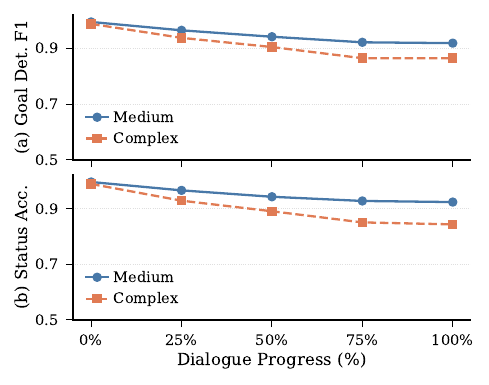}
    \caption{Goal detection F1 (top) and status tracking accuracy (bottom) versus normalized dialogue progress on medium and complex dialogues. The leftmost point is the first available turn-level checkpoint after progress normalization, not an empty-context evaluation.}
    \label{fig:metrics_online}
\end{figure}

\noindent\emph{Online behavior.} Figure~\ref{fig:metrics_online} shows incremental tracking performance as the fixed dialogue is revealed. This evaluates the online replay setting (\S\ref{sec:evaluation_framework}); it does not measure adaptation during interactive agent rollouts.

No baseline dominates both axes. LLM-Rsum detects more goals but trails our complex status tracking and incurs more than $7\times$ the per-turn latency (\S\ref{sec:efficiency}), while RAG's higher medium status accuracy comes with lower detection F1. Our evaluator balances detection and conditional status tracking at the lowest measured per-turn latency among these memory methods.

\subsection{Efficiency Analysis}
\label{sec:efficiency}

\begin{figure}[ht]
    \centering
    \includegraphics[width=\linewidth]{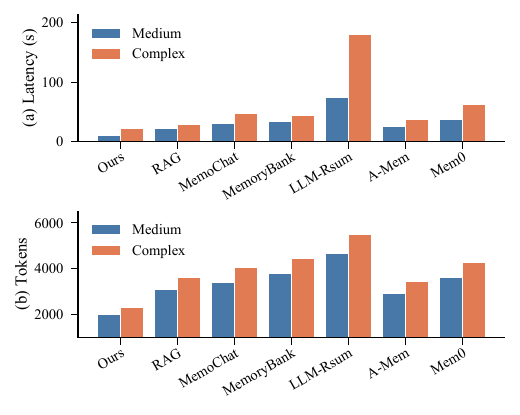}
    \caption{Per-turn update latency (top) and token usage (bottom) across methods.}
    \label{fig:latency_token}
\end{figure}

Figure~\ref{fig:latency_token} reports per-turn latency and token usage. Our evaluator is most efficient on both axes ($<$25s per turn on complex dialogues; 2{,}030/2{,}350 tokens medium/complex). RAG and A-Mem are next (22--38s), Mem0 incurs moderate overhead (38--62s), and LLM-Rsum reaches 180s on complex dialogues with more than twice the tokens because it re-summarizes the full history each turn. Among the evaluated methods, our evaluator lies on the accuracy--latency Pareto frontier.

\subsection{Metric Validity Analysis}

\begin{table}[ht]
\centering
\resulttablestyle
\caption{Average diagnostic scores on fixed synthetic dialogues, not success rates from interactive agent rollouts. Turn-level Relevance is the native 1--5 judge score divided by 5. All other bounded metrics are on a 0--1 scale; Dialogue-level Coherence remains on its native 1--5 scale, and NTC is a turn count.}
\label{tab:metric_results}
\begin{tabular*}{\columnwidth}{@{\extracolsep{\fill}}lcc@{}}
\toprule
\textbf{Metric}                               & \textbf{Medium} & \textbf{Complex} \\
\midrule
dGCR                                            & 0.967           & 0.930 \\
NTC                                             & 7.04            & 10.50 \\
MRA                                             & 0.913           & 0.743 \\
PE                                              & 0.619           & 0.586 \\
Turn-level Relevance                           & 0.752           & 0.766 \\
Dialogue-level Coherence                       & 4.40            & 4.45 \\
\bottomrule
\end{tabular*}
\end{table}

Table~\ref{tab:metric_results} summarizes the fixed dialogues and their reconstructed states. Complex dialogues have lower MRA and larger NTC, while relevance and coherence are slightly higher. We report these metrics separately because they capture different aspects of the dialogues, although these aggregate values alone do not show that the metrics are non-redundant.

\begin{table}[ht]
\centering
\resulttablestyle
\caption{Correlation of proposed metrics with dGCR, reported as Pearson's $r$ and Spearman's $\rho$ under Medium and Complex settings.}
\label{tab:metric_validity}
\begin{tabular*}{\columnwidth}{@{\extracolsep{\fill}}lcccc@{}}
\toprule
\multirow{2}{*}{\textbf{Metric}} & \multicolumn{2}{c}{\textbf{Medium}} & \multicolumn{2}{c}{\textbf{Complex}} \\
\cmidrule(lr){2-3} \cmidrule(lr){4-5}
 & $r$ & $\rho$ & $r$ & $\rho$ \\
\midrule
NTC                               & $+0.08$ & $+0.16$ & $+0.20$ & $+0.05$ \\
MRA                               & $+0.75$ & $+0.60$ & $+0.44$ & $+0.43$ \\
PE                                & $-0.05$ & $-0.03$ & $+0.16$ & $+0.12$ \\
Turn-level Relevance              & $+0.22$ & $+0.29$ & $+0.08$ & $+0.09$ \\
Dialogue-level Coherence          & $-0.11$ & $-0.08$ & $+0.13$ & $+0.25$ \\
\bottomrule
\end{tabular*}
\end{table}

The metrics distinguish different failure modes by construction: two dialogues can end with the same completed and failed goals, and thus the same dGCR, while taking different numbers of turns (NTC) or differing in recall of earlier constraints (MRA). Successful completion also does not establish that unsolicited actions were useful (PE). These distinctions motivate the metric set but are not empirical evidence of independence.

Table~\ref{tab:metric_validity} reports correlations with dGCR as a task-completion anchor. Because dGCR is computed from reconstructed states (\S\ref{sec:task_completion}), these values measure internal alignment rather than independent ground truth. MRA correlates most strongly with dGCR ($r=0.75$ on medium and $0.44$ on complex dialogues), while PE has weak correlations ($r=-0.05$ and $0.16$). Low correlation with dGCR alone does not establish independent information: it may also reflect measurement noise or restricted variation. In particular, the 81.6\% audit agreement for beneficial proactivity limits the interpretation of PE. We use PE as a diagnostic, not a task-success measure or a basis for the main method-ranking claim; we do not claim probabilistic calibration. A full pairwise analysis and incremental prediction of independently measured user success are needed to test redundancy more directly.

\subsection{Validity of LLM Judges}
\label{sec:judge_validity}

\noindent\emph{Manual audit.} We manually reviewed 2{,}880 LLM judge decisions from 100 stratified dialogues. Agreement with this audit is 94.0\% overall (Table~\ref{tab:human_agreement}), but falls to 81.6\% for beneficial proactivity, indicating that this subjective metric should be interpreted cautiously. This audit is an initial check on metric judgments, not an independent multi-annotator validation or manual screening of generated dialogues. The protocol is detailed in Appendix~\ref{appendix:human_agreement}.

\begin{table}[ht]
\centering
\resulttablestyle
\caption{Agreement between LLM judge decisions and a single-author manual audit on 100 stratified dialogues (2{,}880 items). Overall agreement is weighted by item count.}
\label{tab:human_agreement}
\begin{tabular*}{\columnwidth}{@{\extracolsep{\fill}}lcc@{}}
\toprule
\textbf{Metric Dimension} & \textbf{\# Items} & \textbf{Audit Agr.} \\
\midrule
Goal Status Transition   & 1{,}850 & 94.2\% \\
Dependency Relation      & 650     & 96.8\% \\
Proactivity (grounded)   & 190     & 95.3\% \\
Proactivity (beneficial) & 190     & 81.6\% \\
\midrule
\textbf{Overall}         & 2{,}880 & \textbf{94.0\%} \\
\bottomrule
\end{tabular*}
\end{table}

\noindent\emph{Cross-family judge swap.} Replacing the default backbone judge with Qwen3-32B~\cite{yang2025qwen3} shifts our scores by $-1.82$/$-1.71$ on medium F1/Acc and $-3.39$/$+1.02$ on complex. The swap checks whether scores are sensitive to the judge family; it does not rerank all baselines.

\begin{table}[ht]
\centering
\resulttablestyle
\caption{Judge swap for our evaluator (F1 / status accuracy).}
\label{tab:judge_swap}
\begin{tabular*}{\columnwidth}{@{\extracolsep{\fill}}lcccc@{}}
\toprule
\multirow{2}{*}{\textbf{Judge}} & \multicolumn{2}{c}{\textbf{Medium}} & \multicolumn{2}{c}{\textbf{Complex}} \\
\cmidrule(lr){2-3} \cmidrule(lr){4-5}
& F1 & Acc. & F1 & Acc. \\
\midrule
Default LLM Judge & 91.92 & 92.31 & 86.49 & 84.28 \\
Qwen3-32B & 90.10 & 90.60 & 83.10 & 85.30 \\
$\Delta$ & $-$1.82 & $-$1.71 & $-$3.39 & $+$1.02 \\
\bottomrule
\end{tabular*}
\end{table}

\section{Conclusions}
\label{sec:conclusion}
We introduced \textbf{ATOD} and \textbf{ATOD-Eval} for evaluating interleaved, dependency-aware TOD with turn-level lifecycle annotations, together with a symbolic--vector memory evaluator. The evaluator balances detection and tracking on complex dialogues at the lowest measured per-turn latency among matched-backbone baselines. The judge swap and manual audit support the metric judgments, although the audit also shows that beneficial proactivity remains subjective. Future work will extend the benchmark to real interactions and isolate individual memory components.
\section*{Limitations}

\noindent\textbf{Construction and validation.} LLM-based construction and evaluation risk circularity. A verifier pass and author screening check dataset quality; a cross-family judge swap and separate manual audit check metric judgments. Neither provides independent multi-annotator validation.

\noindent\textbf{Uncertainty and attribution.} Point estimates do not quantify dialogue-sampling or API-call variability. Correlations with dGCR do not establish metric independence, and co-occurring properties prevent attribution to individual behaviors. A top-$k$ sweep and same-backbone comparison assess the memory layer, but proactive auditing and the dependency graph lack separate ablations.

\noindent\textbf{Scope.} ATOD comprises held-out synthetic English, text-only dialogues derived from SGD domains. It measures goal-trajectory recovery rather than live user success, without personalization or multimodal inputs.

\section*{Ethical Considerations}

ATOD uses synthetic dialogues generated from public SGD goal structures, not deployed-system conversations. Despite staged quality control, synthetic generation does not guarantee privacy or eliminate model biases. Proactivity can become intrusive or manipulative without user controls, and judgments of benefit vary across cultures and user populations. Downstream evaluation should therefore account for user preferences and these limitations. The dataset is intended solely for research and will be released under the CC BY-NC 4.0 license.


\bibliography{custom}

\begin{thebibliography}{44}
\providecommand{\natexlab}[1]{#1}

\bibitem[{Acikgoz et~al.(2025)Acikgoz, Guo, Dey, Datta, Kim, Tur, and
  Hakkani-T{\"u}r}]{acikgoz2025td}
Emre~Can Acikgoz, Carl Guo, Suvodip Dey, Akul Datta, Takyoung Kim, Gokhan Tur,
  and Dilek Hakkani-T{\"u}r. 2025.
\newblock \href {https://doi.org/10.18653/v1/2025.sigdial-1.7} {{TD}-{EVAL}:
  Revisiting task-oriented dialogue evaluation by combining turn-level
  precision with dialogue-level comparisons}.
\newblock In \emph{Proceedings of the 26th Annual Meeting of the Special
  Interest Group on Discourse and Dialogue}, pages 113--132. Association for
  Computational Linguistics.

\bibitem[{Barres et~al.(2025)Barres, Dong, Ray, Si, and
  Narasimhan}]{barres2025tau2}
Victor Barres, Honghua Dong, Soham Ray, Xujie Si, and Karthik Narasimhan. 2025.
\newblock \href {https://arxiv.org/abs/2506.07982} {{$\tau^2$-Bench}:
  Evaluating conversational agents in a dual-control environment}.
\newblock \emph{arXiv preprint arXiv:2506.07982}.

\bibitem[{Budzianowski et~al.(2018)Budzianowski, Wen, Tseng, Casanueva, Ultes,
  Ramadan, and Ga{\v{s}}i{\'c}}]{budzianowski2018multiwoz}
Pawe{\l} Budzianowski, Tsung-Hsien Wen, Bo-Hsiang Tseng, I{\~n}igo Casanueva,
  Stefan Ultes, Osman Ramadan, and Milica Ga{\v{s}}i{\'c}. 2018.
\newblock \href {https://doi.org/10.18653/v1/D18-1547} {{MultiWOZ} - a
  large-scale multi-domain {Wizard-of-Oz} dataset for task-oriented dialogue
  modelling}.
\newblock In \emph{Proceedings of the 2018 Conference on Empirical Methods in
  Natural Language Processing}, pages 5016--5026. Association for Computational
  Linguistics.

\bibitem[{Chakraborty et~al.(2025)Chakraborty, Dashore, Bathaee, Jain, Das,
  Zhang, Sahu, Naphade, and Winata}]{chakraborty2026t}
Amartya Chakraborty, Paresh Dashore, Nadia Bathaee, Anmol Jain, Anirban Das,
  Shi-Xiong Zhang, Sambit Sahu, Milind Naphade, and Genta~Indra Winata. 2025.
\newblock \href {https://doi.org/10.52202/085713-3479} {{T1}: A tool-oriented
  conversational dataset for multi-turn agentic planning}.
\newblock In \emph{Advances in Neural Information Processing Systems},
  volume~38, pages 115282--115327.

\bibitem[{Chhikara et~al.(2025)Chhikara, Khant, Aryan, Singh, and
  Yadav}]{chhikara2025mem0}
Prateek Chhikara, Dev Khant, Saket Aryan, Taranjeet Singh, and Deshraj Yadav.
  2025.
\newblock \href {https://arxiv.org/abs/2504.19413} {{Mem0}: Building
  production-ready {AI} agents with scalable long-term memory}.
\newblock \emph{arXiv preprint arXiv:2504.19413}.

\bibitem[{Davidson et~al.(2023)Davidson, Romeo, Shu, Gung, Gupta, Mansour, and
  Zhang}]{davidson2023user}
Sam Davidson, Salvatore Romeo, Raphael Shu, James Gung, Arshit Gupta, Saab
  Mansour, and Yi~Zhang. 2023.
\newblock \href {https://arxiv.org/abs/2309.13233} {User simulation with large
  language models for evaluating task-oriented dialogue}.
\newblock \emph{arXiv preprint arXiv:2309.13233}.

\bibitem[{{DeepSeek-AI}(2025)}]{liu2025deepseek}
{DeepSeek-AI}. 2025.
\newblock \href {https://arxiv.org/abs/2512.02556} {{DeepSeek-V3.2}: Pushing
  the frontier of open large language models}.
\newblock \emph{arXiv preprint arXiv:2512.02556}.

\bibitem[{Deng et~al.(2022)Deng, Zhang, Lam, Cheng, and Meng}]{deng2022user}
Yang Deng, Wenxuan Zhang, Wai Lam, Hong Cheng, and Helen Meng. 2022.
\newblock \href {https://doi.org/10.1145/3485447.3512020} {User satisfaction
  estimation with sequential dialogue act modeling in goal-oriented
  conversational systems}.
\newblock In \emph{Proceedings of the ACM Web Conference 2022}, pages
  2998--3008.

\bibitem[{Douze et~al.(2024)Douze, Guzhva, Deng, Johnson, Szilvasy, Mazar{\'e},
  Lomeli, Hosseini, and J{\'e}gou}]{douze2024faiss}
Matthijs Douze, Alexandr Guzhva, Chengqi Deng, Jeff Johnson, Gergely Szilvasy,
  Pierre-Emmanuel Mazar{\'e}, Maria Lomeli, Lucas Hosseini, and Herv{\'e}
  J{\'e}gou. 2024.
\newblock \href {https://arxiv.org/abs/2401.08281} {The {Faiss} library}.
\newblock \emph{arXiv preprint arXiv:2401.08281}.

\bibitem[{Du et~al.(2025)Du, Wang, He, Liang, Wang, Li, Gui, Pan, Xu, and
  Wong}]{du2025bridging}
Yiming Du, Bingbing Wang, Yang He, Bin Liang, Baojun Wang, Zhongyang Li, Lin
  Gui, Jeff~Z. Pan, Ruifeng Xu, and Kam-Fai Wong. 2025.
\newblock \href {https://arxiv.org/abs/2505.20231v1} {Bridging the long-term
  gap: A memory-active policy for multi-session task-oriented dialogue}.
\newblock \emph{arXiv preprint arXiv:2505.20231}.

\bibitem[{Dubois et~al.(2024)Dubois, Galambosi, Liang, and
  Hashimoto}]{dubois2024length}
Yann Dubois, Bal{\'a}zs Galambosi, Percy Liang, and Tatsunori~B. Hashimoto.
  2024.
\newblock \href {https://arxiv.org/abs/2404.04475} {Length-controlled
  {AlpacaEval}: A simple way to debias automatic evaluators}.
\newblock \emph{arXiv preprint arXiv:2404.04475}.

\bibitem[{{Gemini Team}(2025)}]{comanici2025gemini}
{Gemini Team}. 2025.
\newblock \href {https://arxiv.org/abs/2507.06261} {{Gemini 2.5}: Pushing the
  frontier with advanced reasoning, multimodality, long context, and next
  generation agentic capabilities}.
\newblock \emph{arXiv preprint arXiv:2507.06261}.

\bibitem[{{GLM-5 Team}(2026)}]{zeng2026glm}
{GLM-5 Team}. 2026.
\newblock \href {https://arxiv.org/abs/2602.15763} {{GLM-5}: From vibe coding
  to agentic engineering}.
\newblock \emph{arXiv preprint arXiv:2602.15763}.

\bibitem[{Jain et~al.(2025)Jain, Aggarwal, Sahay, Dong, and
  Saladi}]{jain2025autoeval}
Arihant Jain, Purav Aggarwal, Rishav Sahay, Chaosheng Dong, and Anoop Saladi.
  2025.
\newblock \href {https://doi.org/10.18653/v1/2025.naacl-long.508}
  {{AutoEval-ToD}: Automated evaluation of task-oriented dialog systems}.
\newblock In \emph{Proceedings of the 2025 Conference of the Nations of the
  Americas Chapter of the Association for Computational Linguistics: Human
  Language Technologies (Volume 1: Long Papers)}, pages 10133--10148.
  Association for Computational Linguistics.

\bibitem[{Kazi et~al.(2024)Kazi, Lyu, Zhou, Hakkani-T{\"u}r, and
  Tur}]{kazi2024large}
Taaha Kazi, Ruiliang Lyu, Sizhe Zhou, Dilek Hakkani-T{\"u}r, and Gokhan Tur.
  2024.
\newblock \href {https://doi.org/10.1109/SLT61566.2024.10832298} {Large
  language models as user-agents for evaluating task-oriented-dialogue
  systems}.
\newblock In \emph{2024 IEEE Spoken Language Technology Workshop (SLT)}, pages
  913--920. IEEE.

\bibitem[{{Kimi Team}(2026)}]{team2026kimi}
{Kimi Team}. 2026.
\newblock \href {https://arxiv.org/abs/2602.02276} {{Kimi K2.5}: Visual agentic
  intelligence}.
\newblock \emph{arXiv preprint arXiv:2602.02276}.

\bibitem[{Komma et~al.(2023)Komma, Panyam~Chandrasekarasastry, Leffel, Goyal,
  Metallinou, Matsoukas, and Galstyan}]{komma2023toward}
Abishek Komma, Nagesh Panyam~Chandrasekarasastry, Timothy Leffel, Anuj Goyal,
  Angeliki Metallinou, Spyros Matsoukas, and Aram Galstyan. 2023.
\newblock \href {https://doi.org/10.18653/v1/2023.acl-industry.19} {Toward more
  accurate and generalizable evaluation metrics for task-oriented dialogs}.
\newblock In \emph{Proceedings of the 61st Annual Meeting of the Association
  for Computational Linguistics (Volume 5: Industry Track)}, pages 186--195.
  Association for Computational Linguistics.

\bibitem[{Kulkarni et~al.(2024)Kulkarni, Tseng, Moniz, Piraviperumal, Yu, and
  Bhargava}]{kulkarni2024synthdst}
Atharva Kulkarni, Bo-Hsiang Tseng, Joel Ruben~Antony Moniz, Dhivya
  Piraviperumal, Hong Yu, and Shruti Bhargava. 2024.
\newblock \href {https://doi.org/10.18653/v1/2024.eacl-long.120} {{S}ynth{DST}:
  Synthetic data is all you need for few-shot dialog state tracking}.
\newblock In \emph{Proceedings of the 18th Conference of the European Chapter
  of the Association for Computational Linguistics (Volume 1: Long Papers)},
  pages 1988--2001. Association for Computational Linguistics.

\bibitem[{Lewis et~al.(2020)Lewis, Perez, Piktus, Petroni, Karpukhin, Goyal,
  K{\"u}ttler, Lewis, Yih, Rockt{\"a}schel, Riedel, and
  Kiela}]{lewis2020retrieval}
Patrick Lewis, Ethan Perez, Aleksandra Piktus, Fabio Petroni, Vladimir
  Karpukhin, Naman Goyal, Heinrich K{\"u}ttler, Mike Lewis, Wen-tau Yih, Tim
  Rockt{\"a}schel, Sebastian Riedel, and Douwe Kiela. 2020.
\newblock \href
  {https://proceedings.neurips.cc/paper_files/paper/2020/hash/6b493230205f780e1bc26945df7481e5-Abstract.html}
  {Retrieval-augmented generation for knowledge-intensive {NLP} tasks}.
\newblock In \emph{Advances in Neural Information Processing Systems},
  volume~33, pages 9459--9474.

\bibitem[{Li et~al.(2025)Li, Song, Wang, Niu, Chen, Yang, Xi, Lai, Zhao, Wang,
  Ren, Lin, Huo, Chen, Chen, Li, Yin, Yu, Tang, Yang, Xu, and
  Xiong}]{li2025memos}
Zhiyu Li, Shichao Song, Hanyu Wang, Simin Niu, Ding Chen, Jiawei Yang, Chenyang
  Xi, Huayi Lai, Jihao Zhao, Yezhaohui Wang, Junpeng Ren, Zehao Lin, Jiahao
  Huo, Tianyi Chen, Kai Chen, Kehang Li, Zhiqiang Yin, Qingchen Yu, Bo~Tang,
  Hongkang Yang, Zhi-Qin~John Xu, and Feiyu Xiong. 2025.
\newblock \href {https://arxiv.org/abs/2505.22101} {{MemOS}: An operating
  system for memory-augmented generation ({MAG}) in large language models}.
\newblock \emph{arXiv preprint arXiv:2505.22101}.

\bibitem[{Liu et~al.(2024{\natexlab{a}})Liu, Yu, Zhang, Xu, Lei, Lai, Gu, Ding,
  Men, Yang, Zhang, Deng, Zeng, Du, Zhang, Shen, Zhang, Su, Sun, Huang, Dong,
  and Tang}]{liu2024agentbench}
Xiao Liu, Hao Yu, Hanchen Zhang, Yifan Xu, Xuanyu Lei, Hanyu Lai, Yu~Gu,
  Hangliang Ding, Kaiwen Men, Kejuan Yang, Shudan Zhang, Xiang Deng, Aohan
  Zeng, Zhengxiao Du, Chenhui Zhang, Sheng Shen, Tianjun Zhang, Yu~Su, Huan
  Sun, Minlie Huang, Yuxiao Dong, and Jie Tang. 2024{\natexlab{a}}.
\newblock \href {https://openreview.net/forum?id=zAdUB0aCTQ} {{AgentBench}:
  Evaluating {LLM}s as agents}.
\newblock In \emph{The Twelfth International Conference on Learning
  Representations}.

\bibitem[{Liu et~al.(2023)Liu, Iter, Xu, Wang, Xu, and Zhu}]{liu2023g}
Yang Liu, Dan Iter, Yichong Xu, Shuohang Wang, Ruochen Xu, and Chenguang Zhu.
  2023.
\newblock \href {https://doi.org/10.18653/v1/2023.emnlp-main.153} {{G-Eval}:
  {NLG} evaluation using {GPT-4} with better human alignment}.
\newblock In \emph{Proceedings of the 2023 Conference on Empirical Methods in
  Natural Language Processing}, pages 2511--2522. Association for Computational
  Linguistics.

\bibitem[{Liu et~al.(2024{\natexlab{b}})Liu, Fang, Vandyke, and
  Collier}]{liu2024toad}
Yinhong Liu, Yimai Fang, David Vandyke, and Nigel Collier. 2024{\natexlab{b}}.
\newblock \href {https://doi.org/10.18653/v1/2024.findings-acl.494} {{TOAD}:
  Task-oriented automatic dialogs with diverse response styles}.
\newblock In \emph{Findings of the Association for Computational Linguistics:
  ACL 2024}, pages 8341--8356. Association for Computational Linguistics.

\bibitem[{Lu et~al.(2023)Lu, An, Lin, Pergola, He, Yin, Sun, and
  Wu}]{lu2023memochat}
Junru Lu, Siyu An, Mingbao Lin, Gabriele Pergola, Yulan He, Di~Yin, Xing Sun,
  and Yunsheng Wu. 2023.
\newblock \href {https://arxiv.org/abs/2308.08239} {{MemoChat}: Tuning {LLMs}
  to use memos for consistent long-range open-domain conversation}.
\newblock \emph{arXiv preprint arXiv:2308.08239}.

\bibitem[{Maharana et~al.(2024)Maharana, Lee, Tulyakov, Bansal, Barbieri, and
  Fang}]{maharana2024evaluating}
Adyasha Maharana, Dong-Ho Lee, Sergey Tulyakov, Mohit Bansal, Francesco
  Barbieri, and Yuwei Fang. 2024.
\newblock \href {https://doi.org/10.18653/v1/2024.acl-long.747} {Evaluating
  very long-term conversational memory of {LLM} agents}.
\newblock In \emph{Proceedings of the 62nd Annual Meeting of the Association
  for Computational Linguistics (Volume 1: Long Papers)}, pages 13851--13870.
  Association for Computational Linguistics.

\bibitem[{Mo et~al.(2024)Mo, Jiang, Maharaj, Hishamunda, and
  Li}]{mo2024hiertod}
Lingbo Mo, Shun Jiang, Akash Maharaj, Bernard Hishamunda, and Yunyao Li. 2024.
\newblock \href {https://arxiv.org/abs/2411.07152} {{HierTOD}: A task-oriented
  dialogue system driven by hierarchical goals}.
\newblock \emph{arXiv preprint arXiv:2411.07152}.

\bibitem[{Ong et~al.(2025)Ong, Kim, Gwak, Chae, Kwon, Jo, Hwang, Lee, and
  Yeo}]{ong2024towards}
Kai Tzu-iunn Ong, Namyoung Kim, Minju Gwak, Hyungjoo Chae, Taeyoon Kwon, Yohan
  Jo, Seung-won Hwang, Dongha Lee, and Jinyoung Yeo. 2025.
\newblock \href {https://doi.org/10.18653/v1/2025.naacl-long.435} {Towards
  lifelong dialogue agents via timeline-based memory management}.
\newblock In \emph{Proceedings of the 2025 Conference of the Nations of the
  Americas Chapter of the Association for Computational Linguistics: Human
  Language Technologies (Volume 1: Long Papers)}, pages 8631--8661. Association
  for Computational Linguistics.

\bibitem[{{OpenAI}(2025)}]{openai2024chatgptagent}
{OpenAI}. 2025.
\newblock Introducing {ChatGPT} agent: bridging research and action.
\newblock \url{https://openai.com/index/introducing-chatgpt-agent/}.
\newblock Accessed September 11, 2026.

\bibitem[{Packer et~al.(2023)Packer, Wooders, Lin, Fang, Patil, Stoica, and
  Gonzalez}]{packer2023memgpt}
Charles Packer, Sarah Wooders, Kevin Lin, Vivian Fang, Shishir~G. Patil, Ion
  Stoica, and Joseph~E. Gonzalez. 2023.
\newblock \href {https://arxiv.org/abs/2310.08560} {{MemGPT}: Towards {LLMs} as
  operating systems}.
\newblock \emph{arXiv preprint arXiv:2310.08560}.

\bibitem[{Peng et~al.(2021)Peng, Li, Zhang, Zhu, Li, and Gao}]{peng2020raddle}
Baolin Peng, Chunyuan Li, Zhu Zhang, Chenguang Zhu, Jinchao Li, and Jianfeng
  Gao. 2021.
\newblock \href {https://doi.org/10.18653/v1/2021.acl-long.341} {{RADDLE}: An
  evaluation benchmark and analysis platform for robust task-oriented dialog
  systems}.
\newblock In \emph{Proceedings of the 59th Annual Meeting of the Association
  for Computational Linguistics and the 11th International Joint Conference on
  Natural Language Processing (Volume 1: Long Papers)}, pages 4418--4429.
  Association for Computational Linguistics.

\bibitem[{Rana et~al.(2025)Rana, Man, Msiiwa, Paine, Zhu, Dev, Sharma, and
  R}]{rana2025agentchangebench}
Manik Rana, Calissa Man, Anotida~Expected Msiiwa, Jeffrey Paine, Kevin Zhu,
  Sunishchal Dev, Vasu Sharma, and Ahan~M R. 2025.
\newblock \href {https://arxiv.org/abs/2510.18170} {{AgentChangeBench}: A
  multi-dimensional evaluation framework for goal-shift robustness in
  conversational {AI}}.
\newblock \emph{arXiv preprint arXiv:2510.18170}.

\bibitem[{Rastogi et~al.(2020)Rastogi, Zang, Sunkara, Gupta, and
  Khaitan}]{rastogi2020towards}
Abhinav Rastogi, Xiaoxue Zang, Srinivas Sunkara, Raghav Gupta, and Pranav
  Khaitan. 2020.
\newblock \href {https://doi.org/10.1609/aaai.v34i05.6394} {Towards scalable
  multi-domain conversational agents: The schema-guided dialogue dataset}.
\newblock In \emph{Proceedings of the AAAI Conference on Artificial
  Intelligence}, volume~34, pages 8689--8696.

\bibitem[{Sen et~al.(2026)Sen, Lumer, Gulati, and Subbiah}]{sen2026chronos}
Sahil Sen, Elias Lumer, Anmol Gulati, and Vamse~Kumar Subbiah. 2026.
\newblock \href {https://arxiv.org/abs/2603.16862} {{Chronos}: Temporal-aware
  conversational agents with structured event retrieval for long-term memory}.
\newblock \emph{arXiv preprint arXiv:2603.16862}.

\bibitem[{Stacey et~al.(2024)Stacey, Cheng, Torr, Guigue, Driesen, Coca,
  Gaynor, and Johannsen}]{stacey2024lucid}
Joe Stacey, Jianpeng Cheng, John Torr, Tristan Guigue, Joris Driesen, Alexandru
  Coca, Mark Gaynor, and Anders Johannsen. 2024.
\newblock \href {https://doi.org/10.18653/v1/2024.naacl-srw.8} {{LUCID}:
  {LLM}-generated utterances for complex and interesting dialogues}.
\newblock In \emph{Proceedings of the 2024 Conference of the North American
  Chapter of the Association for Computational Linguistics: Human Language
  Technologies (Volume 4: Student Research Workshop)}, pages 56--74.
  Association for Computational Linguistics.

\bibitem[{Walker et~al.(2000)Walker, Kamm, and Litman}]{walker2000towards}
Marilyn Walker, Candace Kamm, and Diane Litman. 2000.
\newblock \href {https://doi.org/10.1017/S1351324900002503} {Towards developing
  general models of usability with {PARADISE}}.
\newblock \emph{Natural Language Engineering}, 6(3-4):363--377.

\bibitem[{Wang et~al.(2023)Wang, Cheng, Lin, Leong, and Li}]{wang2023target}
Jian Wang, Yi~Cheng, Dongding Lin, Chak~Tou Leong, and Wenjie Li. 2023.
\newblock \href {https://doi.org/10.18653/v1/2023.emnlp-main.72}
  {Target-oriented proactive dialogue systems with personalization: Problem
  formulation and dataset curation}.
\newblock In \emph{Proceedings of the 2023 Conference on Empirical Methods in
  Natural Language Processing}, pages 1132--1143. Association for Computational
  Linguistics.

\bibitem[{Wang et~al.(2025)Wang, Fu, Cao, Wang, Tian, and
  Ding}]{wang2025recursively}
Qingyue Wang, Yanhe Fu, Yanan Cao, Shuai Wang, Zhiliang Tian, and Liang Ding.
  2025.
\newblock \href {https://doi.org/10.1016/j.neucom.2025.130193} {Recursively
  summarizing enables long-term dialogue memory in large language models}.
\newblock \emph{Neurocomputing}, 639:130193.

\bibitem[{Wu et~al.(2025)Wu, Zhang, Liang, and Liu}]{wu2025sgmem}
Yaxiong Wu, Yongyue Zhang, Sheng Liang, and Yong Liu. 2025.
\newblock \href {https://arxiv.org/abs/2509.21212} {{SGMem}: Sentence graph
  memory for long-term conversational agents}.
\newblock \emph{arXiv preprint arXiv:2509.21212}.

\bibitem[{Xu et~al.(2025)Xu, Liang, Mei, Gao, Tan, and Zhang}]{xu2025mem}
Wujiang Xu, Zujie Liang, Kai Mei, Hang Gao, Juntao Tan, and Yongfeng Zhang.
  2025.
\newblock \href {https://arxiv.org/abs/2502.12110} {{A-MEM}: Agentic memory for
  {LLM} agents}.
\newblock \emph{arXiv preprint arXiv:2502.12110}.

\bibitem[{Yang et~al.(2025)Yang, Li, Yang, Zhang, Hui, Zheng, Yu, Gao, Huang,
  Lv et~al.}]{yang2025qwen3}
An~Yang, Anfeng Li, Baosong Yang, Beichen Zhang, Binyuan Hui, Bo~Zheng, Bowen
  Yu, Chang Gao, Chengen Huang, Chenxu Lv, et~al. 2025.
\newblock \href {https://arxiv.org/abs/2505.09388} {{Qwen3} technical report}.
\newblock \emph{arXiv preprint arXiv:2505.09388}.

\bibitem[{Yao et~al.(2024)Yao, Shinn, Razavi, and Narasimhan}]{yao2024tau}
Shunyu Yao, Noah Shinn, Pedram Razavi, and Karthik Narasimhan. 2024.
\newblock \href {https://arxiv.org/abs/2406.12045} {{$\tau$-bench}: A benchmark
  for tool-agent-user interaction in real-world domains}.
\newblock \emph{arXiv preprint arXiv:2406.12045}.

\bibitem[{Yoon et~al.(2025)Yoon, Son, So, Kim, Cho, Park, Lee, and
  Kim}]{yoon2025beyond}
Yejin Yoon, Yuri Son, Namyoung So, Minseo Kim, Minsoo Cho, Chanhee Park,
  Seungshin Lee, and Taeuk Kim. 2025.
\newblock \href {https://doi.org/10.18653/v1/2025.emnlp-main.672} {Beyond
  task-oriented and chitchat dialogues: Proactive and transition-aware
  conversational agents}.
\newblock In \emph{Proceedings of the 2025 Conference on Empirical Methods in
  Natural Language Processing}, pages 13280--13306. Association for
  Computational Linguistics.

\bibitem[{Zheng et~al.(2023)Zheng, Chiang, Sheng, Zhuang, Wu, Zhuang, Lin, Li,
  Li, Xing, Zhang, Gonzalez, and Stoica}]{zheng2023judging}
Lianmin Zheng, Wei-Lin Chiang, Ying Sheng, Siyuan Zhuang, Zhanghao Wu, Yonghao
  Zhuang, Zi~Lin, Zhuohan Li, Dacheng Li, Eric~P. Xing, Hao Zhang, Joseph~E.
  Gonzalez, and Ion Stoica. 2023.
\newblock \href {https://doi.org/10.52202/075280-2020} {Judging
  {LLM}-as-a-judge with {MT-Bench} and {Chatbot Arena}}.
\newblock In \emph{Advances in Neural Information Processing Systems},
  volume~36, pages 46595--46623.

\bibitem[{Zhong et~al.(2024)Zhong, Guo, Gao, Ye, and
  Wang}]{zhong2024memorybank}
Wanjun Zhong, Lianghong Guo, Qiqi Gao, He~Ye, and Yanlin Wang. 2024.
\newblock \href {https://doi.org/10.1609/aaai.v38i17.29946} {{MemoryBank}:
  Enhancing large language models with long-term memory}.
\newblock In \emph{Proceedings of the AAAI Conference on Artificial
  Intelligence}, volume~38, pages 19724--19731.

\end{thebibliography}

\appendix

\section{ATOD Dataset: Statistics, Examples, and Construction Details}
\label{appendix:A_datasets}

\subsection{Annotated Dialogue Example}
\label{appendix:dialogue_example}

Table~\ref{tab:dialogue_example} presents a representative medium-complexity dialogue from ATOD, illustrating interleaving, dependency management, and goal abandonment. The dialogue involves four goals across three domains (Calendar, Services, Flights). Goal~2 (salon appointment) depends on Goal~1 (availability check); Goal~3 (SF flight) is abandoned in turn~7 when the user changes plans to Chicago (Goal~4), which is booked while the salon appointment is confirmed. The status column reproduces the released turn-level annotations.

\begin{table*}[ht]
\centering
\caption{Annotated ATOD dialogue excerpt (medium complexity, 4 goals, 3 domains). Status changes shown per turn; only selected turns are displayed, and omitted turns include intermediate system responses. Goals: G1=check availability, G2=book salon (depends on G1), G3=book SF flight, G4=book Chicago flight.}
\label{tab:dialogue_example}
\footnotesize
\setlength{\tabcolsep}{4pt}
\renewcommand{\arraystretch}{1.0}
\begin{tabular}{clp{9.1cm}l}
\toprule
\textbf{Turn} & \textbf{Speaker} & \textbf{Utterance} & \textbf{Status Changes} \\
\midrule
1 & User & I need to check Sarah Johnson's availability on October 12th, between 2 and 5 PM. & G1$\to$\textsc{Open} \\
2 & System & Let me look that up for you right away. & G1$\to$\textsc{Pending} \\
3 & User & While you're checking that, I also need a one-way flight from Boston to San Francisco on June 12th. & G3$\to$\textsc{Open} \\
4 & System & I'll work on both. Sarah has availability from 2--4 PM. I'm looking for morning Delta flights for you. & G3$\to$\textsc{Pending} \\
5 & User & Great---I'd like to book a haircut with her at Bella Salon for 3:30 PM on October 12th. & G2$\to$\textsc{Open} \\
6 & System & I'll book that appointment. Meanwhile, I found several Boston--SF flights. & G2$\to$\textsc{Pending} \\
7 & User & Actually, change the flight to Chicago O'Hare instead, same date. Is there a United flight? & G3$\to$\textsc{Aband.}, G4$\to$\textsc{Open} \\
8 & System & Your salon appointment is confirmed for Oct 12 at 3:30 PM. Searching Chicago flights now. & G2$\to$\textsc{Compl.}, G4$\to$\textsc{Pending} \\
11 & User & Yes, please book that United flight for me. And could you still check on those Delta flights to San Francisco? & G4$\to$\textsc{Compl.} \\
15 & User & No need to book the San Francisco flights, I just needed the information. So I have the salon on Oct 12 and the Chicago flight on June 12? & --- \\
\bottomrule
\end{tabular}
\end{table*}

\subsection{Synthetic Dataset Quality Analysis}
\label{appendix:dataset_quality}

As aggregate evidence for the construction-time verifier, we assess synthetic dialogues along five dimensions using an LLM quality rater following~\citet{liu2023g}. Table~\ref{tab:synthetic_quality} provides automated quality-control scores, complementing the author-performed manual screening described in Appendix~\ref{appendix:quality_control}.

\begin{table}[ht]
\centering
\caption{LLM-based quality ratings on a 1--5 Likert scale.}
\label{tab:synthetic_quality}
\begin{tabular}{lcc}
\toprule
\textbf{Dimension} & \textbf{Medium} & \textbf{Complex} \\
\midrule
Coherence   & 4.04 & 3.92 \\
Fluency     & 5.00 & 5.00 \\
Consistency & 4.42 & 4.04 \\
Relevance   & 4.58 & 4.62 \\
Naturalness & 4.04 & 4.04 \\
\bottomrule
\end{tabular}
\end{table}

\subsection{Goal Extraction, Co-occurrence Graph Statistics, and Sampling Strategy}
\label{appendix:graph_and_sampling}

We first extract goal sequences from the SGD dataset, where each sequence is an ordered list of user goals (domain–intent pairs) within a dialogue. Table~\ref{tab:goal_extraction_stats} summarizes the extraction results. All 10{,}739 sequences are multi-domain, with an average length of 3.90 goals and a range of 2--8 goals. These seed sequences span 16 unique domains and 37 unique intents; the retained ATOD evaluation set covers 36 intents (\S\ref{sec:dataset_construction}).

\begin{table}[ht]
    \small
    \centering
\caption{Summary statistics for extracted goal sequences.}
    \resizebox{0.6\columnwidth}{!}{
    \begin{tabular}{l c}
        \toprule
        \textbf{Statistic}           & \textbf{Value} \\
        \midrule
        Total Goal Sequences         & 10{,}739 \\
        Avg. Sequence Length         & 3.90 \\
        Length Range                 & 2--8 \\
        Unique Domains               & 16 \\
        Unique Intents               & 37 \\
        \bottomrule
    \end{tabular}
    }
    
    \label{tab:goal_extraction_stats}
\end{table}

We then construct a goal co-occurrence graph where each node is a unique goal and edges represent co-occurrence within the same sequence. Table~\ref{tab:goal_graph_stats} shows its statistics. The graph contains 52 nodes and 396 edges, forming a single connected component with relatively high density ($0.2986$) and average degree ($15.23$), indicating frequent goal co-occurrence across dialogues. This structure supports a diverse sampling of multi-goal trajectories, including high-degree hubs (up to 29) and rare goals (degree as low as 2).

\begin{table}[ht]
    \small
    \centering
\caption{Summary statistics for the co-occurrence graph.}
    \resizebox{0.6\columnwidth}{!}{
    \begin{tabular}{l c}
        \toprule
        \textbf{Statistic}           & \textbf{Value} \\
        \midrule
        Total Nodes (Unique Goals)   & 52 \\
        Total Edges (Co-occurrences) & 396 \\
        Graph Density                & 0.2986 \\
        Average Degree               & 15.23 \\
        Max Degree                   & 29 \\
        Min Degree                   & 2 \\
        \bottomrule
    \end{tabular}
    }
    
    \label{tab:goal_graph_stats}
\end{table}

We sample goal trajectories using the stratified random walks described in \S\ref{sec:graph_construction}. Each sampled walk induces a connected subgraph that is then screened against the desired complexity criteria (\S\ref{appendix:complexity_criteria}), ensuring diversity in goal count, domain coverage, and dependency patterns.

\subsection{Complexity Criteria}
\label{appendix:complexity_criteria}

Our pipeline (\S\ref{sec:dataset_construction}) uses a two-category complexity system (\emph{medium} vs. \emph{complex}), combining quantitative thresholds with qualitative LLM analysis for balanced distribution. Table~\ref{tab:complexity_criteria} shows the criteria based on goals, turns, domains, and advanced agentic behaviors.

\begin{table}[ht]
\centering
\small
\caption{Construction-time complexity criteria. Boolean fields use \ding{51} for true/present and \ding{55} for false/absent. ``Failure / obstacle scenario'' corresponds to the \texttt{defectiveness\_label} field in Listing~\ref{lst:annotation_schema}; it denotes an intentional task obstacle or failure outcome, not an annotation defect. Overlapping ranges are resolved by the hybrid classification procedure described below.}
\label{tab:complexity_criteria}
\begin{tabular*}{\columnwidth}{@{\extracolsep{\fill}}lcc@{}}
\toprule
\textbf{Attribute} & \textbf{Medium} & \textbf{Complex} \\
\midrule
Goals & 2--8 & 7+ \\
Turns & 8--35 & 30+ \\
Asynchronous execution & \ding{51} & \ding{51} \\
Interleaving & \ding{51} & \ding{51} \\
Dependencies & $\leq$2 & $\geq$2 \\
Proactivity & \ding{55} & \ding{51} \\
Failure / obstacle scenario & \ding{55} & \ding{51} \\
\bottomrule
\end{tabular*}
\end{table}

For the categorization process, we follow a three-step procedure. First, \emph{goal sampling} draws trajectories under a two-category distribution (default: 65\% medium, 35\% complex; the retained evaluation set contains 428 medium and 572 complex dialogues after classification and quality control). Second, \emph{annotation} enriches sampled goals with slots, dependencies, and realistic characteristics. Third, \emph{hybrid classification} assigns complexity using pre-defined rules combined with LLM analysis, considering quantitative factors (goal count, domain diversity, dependency structures), qualitative factors (goal interdependence and coordination complexity), and realistic dialogue requirements such as interleaving and proactivity needs. These labels are assigned during construction, before evaluator testing, and are shared by every method. The overlapping goal/turn ranges are therefore not sufficient to reconstruct the labels alone; reproducibility requires the construction attributes and saved final label for each dialogue.

\subsection{Annotated Trajectories and Metadata Specification}
\label{sec:annotated_trajectories}
To represent complex goal structures and agentic behaviors in ATOD, we define a formal schema for dialogue trajectories and metadata (\S\ref{sec:annotation_classification}), shown in Listing~\ref{lst:annotation_schema}. The schema captures interleaved goals, slot-filling states, and explicit inter-goal dependencies that arise in advanced TOD settings. Metadata fields encode global dialogue attributes and execution characteristics (e.g., interleaving, proactivity, and asynchronous actions), which are later used to condition dependency-aware and turn-level evaluation signals. Each goal entry specifies its intent, slots, and dependency relations, enabling systematic analysis of goal initiation, suspension, and resumption across multi-goal interactions.

\begin{lstlisting}[style=json, float=tbp, caption={Annotation schema for dialogue trajectories and goal-level metadata.}, label={lst:annotation_schema}]
{
  "dialogue_id": "string",
  "complexity_class": "medium | complex",
  "classification_method": "pre_defined | model_based",
  "dependency_label": "boolean",
  "defectiveness_label": "boolean",
  "metadata": {
    "num_goals": "integer",
    "estimated_turns": "integer",
    "async_execution": "boolean",
    "interleaving": "boolean",
    "proactivity": "boolean"
    },
  "goal_list": [
    {
      "id": "string",
      "domain": "string",
      "intent": "string",
      "slots": ["string", ...],
      "slot_values": {
        "slot_name_1": "value1",
        "slot_name_2": "value2"
      },
      "dependencies": ["goal_id", ...],
      "content": "string",
      "core_content": "string"
    }
    // ...more goals
  ]
}
\end{lstlisting}

\subsection{ATOD: Quality Control}
\label{appendix:quality_control}
We apply layered screening during construction. Claude-Sonnet-4 performs trajectory annotation, dialogue generation, turn-level status annotation, and both verification stages. The stages use separate calls, but all calls use the same model family and are not independent. Before dialogue generation, a verifier judge checks each annotated trajectory for clear goals, plausible slots, complete fields, and consistent dependencies. After dialogue generation and status annotation, another verifier checks whether the dialogue follows its trajectory, whether lifecycle annotations agree with the utterances, and whether the interaction is coherent and plausible. Flagged instances are retried or filtered. An author then screens the generated dialogues and status annotations for the same failure types. Because this screening was not independently replicated, we treat it as quality control rather than an inter-annotator validation study. The pre-generation verifier prompt is shown below.

\begin{table}[ht]
\small
\centering
\caption{Prompt template for quality assessment of annotated trajectories.}
\begin{tabularx}{\columnwidth}{>{\raggedright\arraybackslash}X}
\toprule
\textbf{Quality Assessment Prompt} \\ \midrule
\texttt{You are a quality judge for annotated goal trajectories.} \\[0.3em]
\textbf{Input:} TRAJECTORY (\{num\_goals\} goals, \{complexity\} complexity): \{goals\_text\} \\[0.3em]
\textbf{Task:} Assess whether this trajectory is ready for dialogue generation. Check: \\
- Goal descriptions are clear and specific \quad - Slot values are realistic (no placeholders) \\
- All required fields are present \quad - Annotations are logically consistent \\[0.3em]
\textbf{Output format:} Respond with exactly one word: \texttt{PASS} or \texttt{FAIL} \\
\bottomrule
\end{tabularx}

\end{table}

\subsection{Dialogue Generation Prompt}
\label{appendix:synthetic_gen_annotation_prompt}

Table~\ref{tab:dialogue_generation_prompt} shows the prompt used for LLM-based dialogue generation in \S\ref{sec:dialog_generation}. Placeholders are filled from the annotated trajectory metadata (\S\ref{sec:annotated_trajectories}).

\begin{table}[ht]
\small
\centering
\caption{Prompt template for dialogue generation.}
\label{tab:dialogue_generation_prompt}
\begin{tabularx}{\columnwidth}{>{\raggedright\arraybackslash}X}
\toprule
\textbf{Dialogue Generation Prompt Template} \\ \midrule
\texttt{Generate a realistic task-oriented dialogue between USER and SYSTEM.} \\[0.3em]
\textbf{Requirements:} Complexity: \{complexity\} \quad Length: \{estimated\_turns\} turns \quad Goals: \{goal\_descriptions\} \quad Attributes: \{agentic\_attrs\} \\[0.3em]
\{combined\_guidance\} \quad \{outcome\_guidance\} \\[0.3em]
\textbf{Dialogue Structure:} (1) User introduces goals naturally; (2) System works under realistic constraints; (3) Natural obstacles and preference changes may occur; (4) Dialogue ends at a natural stopping point; (5) Goal completion may be partial or conditional. \\[0.3em]
\textbf{Natural Conversation Patterns:} Users express needs as they arise; system handles practical constraints; users may add, revise, or abandon goals; availability or technical limitations may surface. \\[0.3em]
\textbf{Format:} Alternating USER/SYSTEM turns, starting with USER. \\
\bottomrule
\end{tabularx}

\end{table}

\subsection{Goal Status Annotation Prompt}
\label{appendix:status_annotation_prompt}

Table~\ref{tab:status_annotation_prompt} shows the prompt used for turn-level goal status annotation in \S\ref{sec:status_annotation}. Listing~\ref{lst:annotated_dialogue} provides a sample annotated dialogue.

\begin{table}[ht]
\small
\centering
\caption{Prompt template for turn-level goal status annotation.}
\label{tab:status_annotation_prompt}
\begin{tabularx}{\columnwidth}{>{\raggedright\arraybackslash}X}
\toprule
\textbf{Goal Status Annotation Prompt Template} \\ \midrule
\texttt{You are tracking goal status in a task-oriented dialogue. Analyze only the current turn and update statuses based on what actually happens.} \\[0.3em]
\textbf{Current Turn:} \{last\_turn\} \quad \textbf{Goals and Current Statuses:} \{goal\_descriptions\} \\[0.3em]
\textbf{Status Meanings:} \\
- \texttt{NOT\_MENTIONED}: not yet appeared \quad - \texttt{OPEN}: mentioned, no action \quad - \texttt{PENDING}: system working \\
- \texttt{COMPLETED}: successfully finished \quad - \texttt{FAILED}: failed \quad - \texttt{ABANDONED}: user cancelled \\[0.3em]
\textbf{Critical Rules:} (1) Change status only if something definitive occurs; (2) \texttt{PENDING} $\to$ only \texttt{COMPLETED}/\texttt{FAILED}/\texttt{ABANDONED}; (3) Preserve status if no clear change. \\[0.3em]
\textbf{Terminal States:} \{goal\_id: current\_status, \dots\} \quad \textbf{JSON Template:} \{json\_template\} \\[0.3em]
\textbf{Instruction:} Respond with only the JSON, updating only goals whose status clearly changes. \\
\bottomrule
\end{tabularx}

\end{table}

\begin{lstlisting}[style=json, float=tbp, caption={Sample annotated ATOD dialogue with turn-level status tracking.}, label={lst:annotated_dialogue}]
{
  "dialogue_id": "...",
  "complexity_class": "complex",
  "metadata": {
    "num_goals": ...,
    "num_turns": ...,
    "async_execution": true,
    "interleaving": true,
    "proactivity": true
  },
  "goal_list": [...],
  "turns": [
    {
      "turn_id": 1,
      "speaker": "USER",
      "utterance": "I need to book a hotel in Chicago.",
      "goal_status_changes": [
        {"goal_id": "g1", "new_status": "open"}
      ],
      "all_goals": {
        "g1": "open",
        "g2": "not_mentioned",
        "g3": "not_mentioned"
      }
    }
    // ... remaining turns omitted
  ]
}
\end{lstlisting}

\section{Robustness and Sensitivity Analyses}
\label{appendix:robustness}

This appendix details the single-author manual audit of metric judgments and top-$k$ retrieval sensitivity. The cross-family judge-swap results are reported in \S\ref{sec:judge_validity}. All analyses use the same 1{,}000-dialogue test set unless otherwise noted.

\subsection{Single-Author Manual Audit of Metric Judgments}
\label{appendix:human_agreement}

\noindent\textbf{Protocol.} We sample a stratified subset of 100 dialogues (50 medium, 50 complex) and extract 2{,}880 individual judge decisions across four dimensions: goal status transitions, dependency relations, proactivity grounding, and proactivity benefit. Under resource constraints, one author manually inspects each decision against the dialogue context, reference annotation, and metric definition, marking whether the LLM judgment is acceptable. Table~\ref{tab:human_agreement_appendix} reports agreement between the LLM decisions and this manual audit. Since no independent second annotator or adjudication was available, we do not report inter-annotator agreement and interpret the audit as a preliminary validity check.

\begin{table}[H]
\centering
\caption{Agreement between LLM judge decisions and a single-author manual audit on 100 stratified dialogues (2{,}880 items). Overall agreement is weighted by item count.}
\label{tab:human_agreement_appendix}
\resizebox{0.7\columnwidth}{!}{
\begin{tabular}{lcc}
\toprule
\textbf{Metric} & \textbf{\# Items} & \textbf{Audit Agr.} \\
\midrule
Goal Status Transition   & 1{,}850 & 94.2\% \\
Dependency Relation      & 650 & 96.8\% \\
Proactivity (grounded)   & 190 & 95.3\% \\
Proactivity (beneficial) & 190 & 81.6\% \\
\midrule
Overall                  & 2{,}880 & 94.0\% \\
\bottomrule
\end{tabular}
}
\end{table}

\noindent\textbf{Findings.} The single-author audit agrees with the LLM judge on 94.0\% of items overall. Agreement is highest for dependency relations (96.8\%), while proactivity \emph{benefit} is lowest (81.6\%), consistent with the subjectivity of that dimension. These results provide an initial error check for the metric judges; they are distinct from the construction-time screening in Appendix~\ref{appendix:quality_control} and are not independent multi-annotator validation.

\subsection{Top-\texorpdfstring{$k$}{k} Retrieval Sensitivity}
\label{appendix:topk_sensitivity}
The top-$k$ retrieval hyperparameter controls how many neighbors are considered during existence checking (Stage 1 in \S\ref{sec:turn_processing}). We sweep $k \in \{1, 3, 5, 7, 9\}$ to assess sensitivity. All other settings remain at their default values (Table~\ref{tab:implementation_hyperparams}).

\begin{figure}[H]
\centering
\includegraphics[width=\columnwidth]{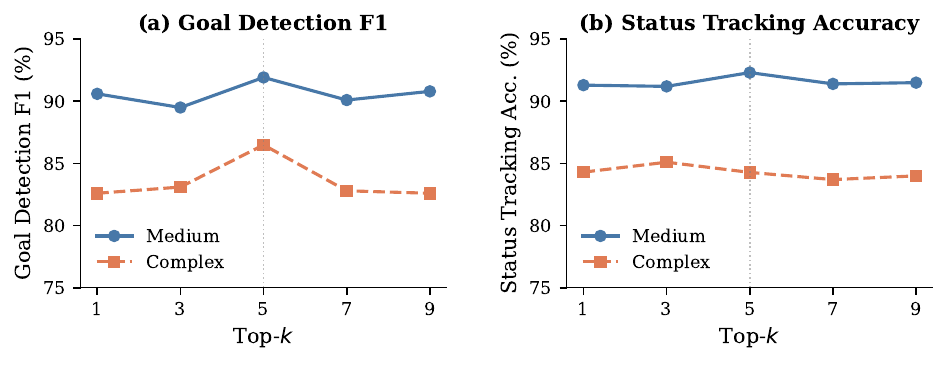}
\caption{Top-$k$ retrieval sensitivity. Goal detection F1 (left) and status tracking accuracy (right) as functions of top-$k$ on the 1{,}000-dialogue test set. The default $k{=}5$ is marked by the dotted vertical line.}
\label{fig:topk_sensitivity}
\end{figure}

On medium dialogues, goal detection F1 varies by less than 2.5 percentage points across the tested values of $k$; conditional status accuracy varies by less than 1.5 points in each split. The default $k{=}5$ achieves the best overall performance, particularly on complex dialogues where it outperforms other settings by 3--4 points in goal F1. Smaller values ($k{=}1, 3$) risk missing relevant neighbors during existence checking, while larger values ($k{=}7, 9$) introduce noise from less relevant candidates. The larger variation in complex-dialogue F1 indicates greater sensitivity to retrieval breadth in that split.

\section{Evaluator Implementation and Prompt Templates}
\label{appendix:memory_system}

\subsection{Implementation Details} 
\label{appendix:memory_implement_details}
The memory system uses Claude-Sonnet-4 (accessed via the Amazon Bedrock API) for goal extraction, existence checking, relationship analysis, and proactive status auditing. Retrieval uses \texttt{all-MiniLM-L6-v2} sentence embeddings (384 dimensions) indexed with FAISS. An LLM classifies links among the top-$k$ retrieved goals rather than relying on a fixed similarity cutoff. Table~\ref{tab:implementation_hyperparams} lists the settings used for all reported results.

The default memory-based runs use Claude-Sonnet-4 via the Amazon Bedrock API (temperature $0$, seed $42$). Every memory-based baseline (Mem0, MemoChat, MemoryBank, RAG, A-Mem, LLM-Rsum) uses the same backbone LLM and \texttt{all-MiniLM-L6-v2} embedding model as our evaluator. The prompt-only baselines in Table~\ref{tab:main_results} are the LLMs under test. This setup controls for model capacity within the memory-based block. We examine potential circularity between the evaluator and metric judges through a Qwen3-32B judge swap and the manual audit in \S\ref{sec:judge_validity}; Appendix~\ref{appendix:human_agreement} gives the audit protocol. FAISS retrieval runs on a single NVIDIA A10 GPU, while LLM calls account for most of the end-to-end latency.

Both hyperparameters were selected on a small held-out development set of 50 dialogues drawn from the ATOD generation pipeline using the same procedure as the test split, then frozen before the 1,000-dialogue evaluation. The existence-match threshold $\tau{=}0.40$ was tuned to favor recall over precision during goal matching: a lower threshold ensures semantically related but differently phrased goals are linked, with the subsequent LLM verification step filtering false positives. Top-$k{=}5$ retrieval balances coverage against prompt length; most dialogues in the development set contained fewer than 10 concurrent goals, so $k{=}5$ retrieves the majority of relevant candidates without exceeding context limits. Sensitivity to $k$ on the full test set is reported in Appendix~\ref{appendix:topk_sensitivity}.

\begin{table}[htbp]
\centering
\caption{Implementation settings for the agentic memory system.}
\label{tab:implementation_hyperparams}
\resizebox{0.85\columnwidth}{!}{
\begin{tabular}{lc}
\toprule
\textbf{Setting} & \textbf{Value} \\
\midrule
Primary LLM Judge & Backbone LLM (see text) \\
Embedding Model & \texttt{all-MiniLM-L6-v2} (384-d) \\
Vector Store / Index Type & FAISS \texttt{IndexFlatIP} \\
Similarity Metric & Cosine (IP on L2-normalized) \\
Top-$k$ Retrieval & 5 \\
Existence-Match Threshold $\tau$ & 0.40 \\
Proactive Auditing Frequency & Every turn \\
\bottomrule
\end{tabular}
}
\end{table}

\subsection{Agentic Memory System Templates}
\label{appendix:memory_system_template}
The memory system uses three LLM prompt templates: goal extraction from individual turns, turn-level status classification, and goal-graph updates for inter-goal links and dependencies. The prompts list five active labels. The sixth state, \textsc{Not\_Mentioned}, is assigned implicitly to reference goals that have not yet appeared and does not require an LLM call.

\begin{table}[ht]
\small
\centering
\caption{Prompt template for goal extraction in the agentic memory system.}
\begin{tabularx}{\columnwidth}{>{\raggedright\arraybackslash}X}
\toprule
\textbf{Goal Extraction Prompt} \\ \midrule
\texttt{Extract user goals from this conversation turn. Use standardized core\_content patterns.} \\[0.3em]
\textbf{Conversation Turn:} \\
User: \{user\_utterance\} \\
System: \{system\_response\} \\[0.3em]
\textbf{Core Content Patterns (examples):} \\
- ``book hotel'' --- hotels, rentals \quad - ``book flight'' --- flights \quad - ``book ticket'' --- bus, concert, train \\
- ``check account'' --- balance, account info \quad - ``search restaurant'' --- find restaurants \\[0.3em]
\textbf{Status Labels:} \texttt{OPEN}, \texttt{PENDING}, \texttt{COMPLETED}, \texttt{FAILED}, \texttt{ABANDONED} \\[0.3em]
\textbf{Output format (JSON array):} \\
\texttt{[\{"goal\_content": "...", "core\_content": "...", "status": "OPEN"\}, ...]} \\
\bottomrule
\end{tabularx}

\label{tab:goal_extraction_prompt}
\end{table}

\begin{table}[ht]
\small
\centering
\caption{Prompt template for per-turn goal status classification.}
\begin{tabularx}{\columnwidth}{>{\raggedright\arraybackslash}X}
\toprule
\textbf{Goal Status Classification Prompt} \\ \midrule
\texttt{Analyze this conversation turn and classify the status of the specific goal below.} \\[0.3em]
\textbf{Goal to classify:} ``\{goal\_content\}'' \\[0.3em]
\textbf{Conversation Turn:} \\
User: \{user\_utterance\} \quad System: \{system\_response\} \\[0.3em]
\textbf{Status Definitions:} \\
- \texttt{OPEN}: mentioned, no action yet \quad - \texttt{PENDING}: system processing \\
- \texttt{COMPLETED}: successfully achieved \quad - \texttt{FAILED}: explicitly failed \quad - \texttt{ABANDONED}: cancelled by user \\[0.3em]
\textbf{Transition Examples:} \\
``book a flight'' $\to$ OPEN \quad ``which dates?'' $\to$ PENDING \quad ``flight booked'' $\to$ COMPLETED \\
``no flights available'' $\to$ FAILED \quad ``never mind'' $\to$ ABANDONED \\[0.3em]
\textbf{Output format (JSON):} \texttt{\{"status": "STATUS"\}} \\
\bottomrule
\end{tabularx}

\label{tab:status_classification_prompt}
\end{table}

\begin{table}[ht]
\small
\centering
\caption{Prompt template for goal graph evolution.}
\begin{tabularx}{\columnwidth}{>{\raggedright\arraybackslash}X}
\toprule
\textbf{Goal Evolution Prompt} \\ \midrule
\texttt{Analyze relationships between a new goal and existing related goals.} \\[0.3em]
\textbf{New Goal:} Content: \{new\_goal.content\} \quad Core Content: \{new\_goal.core\_content\} \\[0.3em]
\textbf{Related Goals (top-$k$ by semantic similarity):} \{related\_goals\_context\} \\[0.3em]
\textbf{Task:} For each related goal, determine the relationship type: \\
- \texttt{"link"}: semantically related but independent \\
- \texttt{"dependency"}: new goal depends on the related goal \\
- \texttt{"none"}: no significant relationship \\[0.3em]
\textbf{Output format (JSON):}\par
\texttt{\{"goal\_id\_1": "relationship\_type", "goal\_id\_2": "relationship\_type"\}} \\
\bottomrule
\end{tabularx}

\label{tab:evolve_prompt}
\end{table}

\subsection{Evaluation Prompt Templates}
\label{appendix:evaluation_prompt_templates}
To make the evaluation protocol explicit, we also provide the prompt templates used for LLM-based judgment of memory recall, proactive behavior, and response quality. In all cases, the judge is given the dialogue context, the system output under evaluation, and the corresponding benchmark reference derived from ATOD annotations.

\begin{table}[htbp]
\small
\centering
\caption{Prompt template for Memory Recall Accuracy evaluation.}
\begin{tabularx}{\columnwidth}{>{\raggedright\arraybackslash}X}
\toprule
\textbf{Memory Recall Accuracy Prompt} \\ \midrule
\texttt{You are evaluating whether a retrieved memory output correctly matches the benchmark memory state.} \\[0.3em]
\textbf{Dialogue Context:} \{dialogue\_context\} \\
\textbf{Retrieval Query:} \{memory\_query\} \\
\textbf{Ground-Truth Memory State:} \{gold\_memory\_json\} \\
\textbf{Retrieved Memory Output:} \{predicted\_memory\} \\[0.3em]
\textbf{Judging Criteria:} \\
- Retrieved output contains information required to answer the query \\
- Goal statuses, slot values, and historical facts are semantically consistent with gold state \\
- No contradiction is introduced \\[0.3em]
\textbf{Output format (JSON):} \texttt{\{"match": 0 or 1\}} \\
\bottomrule
\end{tabularx}

\end{table}

\begin{table}[htbp]
\small
\centering
\caption{Prompt template for Proactivity Effectiveness evaluation.}
\begin{tabularx}{\columnwidth}{>{\raggedright\arraybackslash}X}
\toprule
\textbf{Proactivity Effectiveness Prompt} \\ \midrule
\texttt{You are evaluating whether a system's proactive action is contextually grounded and genuinely helpful.} \\[0.3em]
\textbf{Dialogue Context:} \{dialogue\_context\} \\
\textbf{Current System Action:} \{system\_action\} \\
\textbf{Relevant Goal State:} \{goal\_state\_json\} \\[0.3em]
\textbf{Judging Criteria:} \\
- The action is not explicitly requested by the user in the current turn \\
- The action is grounded in prior dialogue and memory state \\
- The action advances an unresolved or dependency-unlocked goal, or provides a timely reminder \\
- The action is not irrelevant, redundant, or distracting \\[0.3em]
\textbf{Output format (JSON):}\par
\texttt{\{"grounded": 0 or 1, "beneficial": 0 or 1\}} \\
\bottomrule
\end{tabularx}

\end{table}

\begin{table}[htbp]
\small
\centering
\caption{Prompt template for Response Quality evaluation.}
\begin{tabularx}{\columnwidth}{>{\raggedright\arraybackslash}X}
\toprule
\textbf{Response Quality Prompt} \\ \midrule
\texttt{You are evaluating the response quality of a task-oriented dialogue system.} \\[0.3em]
\textbf{Dialogue Context:} \{dialogue\_context\} \\
\textbf{Response Under Evaluation:} \{system\_response\} \\
\textbf{Goal State / Dialogue Reference:} \{goal\_state\_or\_full\_dialogue\} \\[0.3em]
\textbf{Turn-Level Relevance:} addresses current request; consistent with tracked goals; useful and natural \\
\textbf{Dialogue-Level Coherence:} globally consistent; interleaved goals progress coherently; no contradictions \\[0.3em]
\textbf{Output format (JSON):}\par
\texttt{\{"turn\_quality": 1--5, "dialogue\_quality": 1--5\}} \\
\bottomrule
\end{tabularx}

\end{table}

\end{document}